\pdfoutput=1  % arXiv: compile with pdflatex
\documentclass[letterpaper]{article} % DO NOT CHANGE THIS
\usepackage{aaai2027}  % DO NOT CHANGE THIS (fonts are loaded by the style: newtxtext, helvet, courier)
\usepackage[hyphens]{url}  % DO NOT CHANGE THIS
\usepackage{graphicx} % DO NOT CHANGE THIS
\usepackage{natbib}  % DO NOT CHANGE THIS AND DO NOT ADD ANY OPTIONS TO IT
\usepackage{caption} % DO NOT CHANGE THIS AND DO NOT ADD ANY OPTIONS TO IT
\usepackage{amsmath,amssymb}
\usepackage{booktabs}
\usepackage{array}
\usepackage{colortbl}
\usepackage{xcolor}
\usepackage[ruled,vlined,linesnumbered]{algorithm2e}
\usepackage{tikz}
\usetikzlibrary{arrows.meta,positioning,fit,backgrounds,calc,shapes.misc}
\usepackage{newfloat}
\usepackage{listings}
\usepackage{fancyhdr}
\DeclareCaptionStyle{ruled}{labelfont=normalfont,labelsep=colon,strut=off}
\definecolor{ours}{HTML}{C2410C}      % Deep Terracotta / Rust: SARBLOH
\definecolor{base}{HTML}{2563EB}      % Slate Blue: Duck Baseline
\definecolor{ink}{HTML}{0F172A}       % Slate 900: Primary Text
\definecolor{muted}{HTML}{64748B}     % Slate 500: Neutral Muted
\definecolor{paper}{HTML}{FAF7F0}     % Alabaster Paper
\definecolor{oursTint}{HTML}{ECEEFF}  % table highlight for our column (pale periwinkle)
\definecolor{indigo}{HTML}{5B50E6}    % table accent: hatched indigo
\definecolor{periwinkle}{HTML}{939BFF}
\definecolor{lavender}{HTML}{A090FF}
\definecolor{gridc}{HTML}{D3D3D6}     % faint table rules
\definecolor{soft}{HTML}{FFF7ED}      % Soft Orange Tint
\definecolor{softb}{HTML}{EFF6FF}     % Soft Blue Tint

\definecolor{linkc}{HTML}{3B33B8}     % links and caption labels (dark indigo)

\usepackage[most]{tcolorbox}
\newtcolorbox{tintbox}{colback=indigo!4, colframe=indigo!18, boxrule=0.4pt, arc=3pt,
  left=6pt, right=6pt, top=5pt, bottom=5pt}
\newtcblisting{briefing}[1]{listing only, breakable, colback=indigo!3, colframe=indigo!22, boxrule=0.4pt, arc=2pt,
  left=4pt, right=4pt, top=2pt, bottom=2pt, title={#1}, fonttitle=\small\bfseries, coltitle=ink,
  colbacktitle=indigo!10, listing options={basicstyle=\scriptsize\ttfamily, breaklines=true,
  columns=fullflexible, keepspaces=true, showstringspaces=false}}
\tcbset{budgetbox/.style={width=0.495\textwidth, nobeforeafter, box align=top, equal height group=budget,
  colback=white, colframe=indigo!22, boxrule=0.4pt, arc=2pt, left=4pt, right=4pt, top=2pt, bottom=2pt,
  fonttitle=\small\bfseries, coltitle=ink, colbacktitle=indigo!10,
  fontupper=\fontsize{5.6}{6.6}\selectfont\ttfamily}}
\newcommand{\bmark}[1]{\textcolor{#1}{\rule[0.05ex]{0.7ex}{1.2ex}}\,}

\DeclareCaptionLabelSeparator{bar}{\;\textbar\;}
\makeatletter
\let\sarb@PackageError\PackageError
\def\sarb@aaai{aaai}
\def\PackageError#1#2#3{\def\sarb@pkg{#1}%
  \ifx\sarb@pkg\sarb@aaai\PackageWarning{aaai}{#2}\else\sarb@PackageError{#1}{#2}{#3}\fi}
\AtBeginDocument{\AtBeginDocument{\let\PackageError\sarb@PackageError}}
\makeatother
\usepackage[hyphens]{xurl}
\usepackage[colorlinks=true,linkcolor=linkc,citecolor=linkc,urlcolor=linkc,
  pdftitle={Transforming Integer Grids to Spatial Intuition: Perception, Not Planning, Governs ARC-AGI-3},
  pdfauthor={Harsidak Singh Banwait}]{hyperref}

\newcommand{\code}[1]{\texttt{#1}}
\newcolumntype{L}[1]{>{\raggedright\arraybackslash}p{#1}}

\SetCommentSty{algcomment}
\SetKwComment{Comment}{$\triangleright$\ }{}
\SetKwProg{Fn}{function}{}{end}
\SetAlFnt{\small}
\SetAlCapFnt{\small}
\DontPrintSemicolon
\SetArgSty{textup}  % conditions in upright text, like the rest of each line
\SetKwComment{StageComment}{}{}
\newcommand{\Stage}[1]{\StageComment{\upshape\bfseries\textcolor{linkc}{#1}}}  % stage heading: a comment, so unnumbered

\nocopyright  % no AAAI copyright slug on page 1; the running footer below carries the label

\fancypagestyle{plain}{%
  \fancyhf{}%
  \setlength{\footskip}{20pt}%
  \fancyfoot[C]{\footnotesize\textcolor{muted}{arXiv preprint}}%
}
\title{Transforming Integer Grids to Spatial Intuition:\\Perception, Not Planning, Governs ARC-AGI-3}
\author{
    Harsidak Singh Banwait
}
\affiliations{
    Independent Researcher\\
    \texttt{harsidak.singh.banwait.13@gmail.com}
}

\begin{document}

\maketitle
\thispagestyle{plain}

\begin{abstract}
\begin{tintbox}
Sample-efficient fluid intelligence requires autonomous agents that rapidly form operational models of unfamiliar interactive environments. ARC-AGI-3 evaluates this ability across uninstructed visual games under Relative Human Action Efficiency, penalizing exploratory waste quadratically. When frontier reasoning models interact with raw integer grids, discrete serialization destroys spatial topology, inducing perceptual misinduction where models squander hundreds of actions rediscovering elementary boundaries and symmetries. We introduce \textsc{Sarbloh}, an autonomous open-weight agent architecture structured on the premise that spatial perception, rather than planning capacity, governs interactive reasoning on ARC-AGI-3. A zero-parameter, $\mathcal{O}(HW)$ perception engine reconstructs frames in 35\,ms into an object inventory with bipartite temporal tracking, a geometric briefing separating proven invariants from affordance hypotheses under the Measured--Guessed Rule, and an annotated spatial canvas. Paired with a speculative code-as-action runtime, the agent synthesizes algorithmic hypotheses rather than low-level motor primitives. Across eight structural presentations, grounded perception elevates benchmark rubric acquisition from $3/28$ to $25/28$ points ($8.3\times$) while reducing token consumption by $89.4\%$. On twenty-three held-out games, geometric invariants generalize unconditionally with $100\%$ spatial accuracy, whereas single-frame affordance guesses drop to between $6\%$ and $50\%$, confirming the necessity of active hypothesis verification probes. In public evaluation, Run~A (briefing off; it characterizes the action interface, serving loop, and scheduler) resolved 143 of 183 levels in 8.2 hours with a median human action ratio of $0.77\times$ and $59.4\%$ single-pass efficiency, with error decompositions attributing seventy-two percent of unearned score loss to perceptual misinduction on stalled environments; Run~B (briefing on) recorded $0.68\times$ and $63.5\%$ single-pass efficiency. Within the official nine-hour budget on a single Kaggle NVIDIA RTX PRO 6000 GPU, \textsc{Sarbloh} evaluated on the held-out benchmark with the full perception engine active, achieving an official hidden Kaggle score of 16.53, advancing $13.7\times$ beyond the first Milestone winner.
\end{tintbox}
\end{abstract}

\vspace{0.2em}
\noindent\textbf{Code:} \url{https://github.com/Harsidak/SARBLOH}
\vspace{0.35em}

% Headline empirical teaser figure placed after abstract (BAPO-style empirical teaser)
\begin{figure*}[t]
\centering
\includegraphics[width=\textwidth]{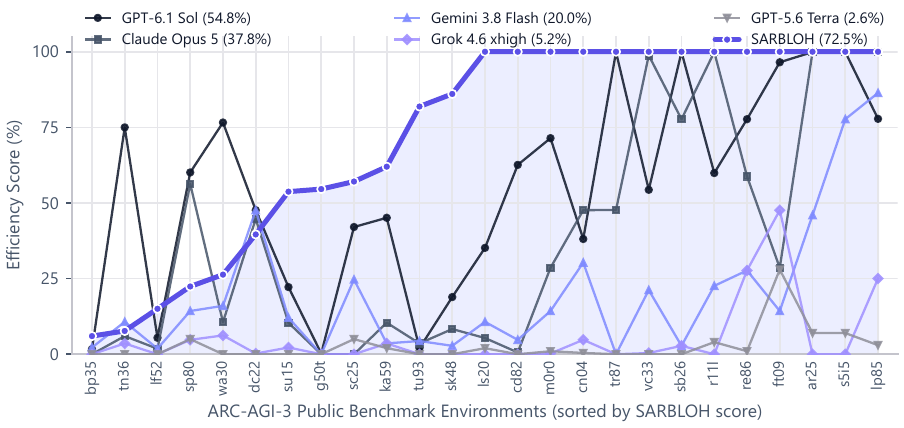}
\caption{\textbf{Action efficiency across 25 public ARC-AGI-3 environments.} 
\textsc{Sarbloh} (open-weight \texttt{Qwen3.8-Flash-Next} on one GPU) leads five frontier API baselines on 21 of 25 games (72.5\% peak mean vs.\ 59.4\% single-pass mean; 100\% on 13 games). Because the strict 9-hour Kaggle ceiling precludes exhausting all 25 games in one pass, the curve reports the best score per game across runs, including isolated re-evaluations on complex environments (\texttt{bp35}).}
\label{fig:pergame}
\end{figure*}

\section{Introduction}
\label{sec:intro}

\textbf{Autonomous skill acquisition under epistemic uncertainty represents a foundational benchmark for artificial general intelligence.} When human players encounter an unfamiliar interactive environment, they do not read procedural manuals---they visually parse the scene within hundreds of milliseconds, effortlessly abstracting embodiment, spatial boundaries, and functional affordances from core-knowledge visual priors \citep{spelke2007core}. The ARC-AGI-3 benchmark \citep{arcagi3} formalizes this capability by evaluating agents across uninstructed, sequential game environments using Relative Human Action Efficiency (RHAE)---a metric that penalizes superfluous exploratory actions quadratically \citep{chollet2019measure}. Under the strict rules of the official offline track, candidate agents must resolve undisclosed games within a nine-hour wall-clock limit on a single NVIDIA RTX PRO 6000 GPU without external network connectivity, precluding both human intervention and unconstrained exploratory rollouts.

\textbf{However, deploying language and vision-language reasoning models in this interactive regime reveals a severe representational bottleneck.} While modern reasoning backbones demonstrate strong symbolic deduction when provided with clean declarative state descriptions, their execution collapses when exposed to raw $64{\times}64$ integer coordinate matrices. In preliminary experiments (Section~\ref{sec:motivation}), a state-of-the-art frontier model expended over 200--300 exploratory actions navigating an elementary maze serialized as raw tokens, exhausting its action budget and failing the task. In stark contrast, providing the identical model with an aligned visual rendering and geometric breakdown reduced required actions to 6--12, surpassing human baselines. This divergence demonstrates that policy deduction is rarely the primary failure mode; rather, unparsed token grids systematically destroy topological adjacency, forcing autoregressive attention to reconstruct basic spatial relations at prohibitive token and action costs.

\textbf{To resolve this fundamental bottleneck, we conduct extensive empirical ablations and establish two central insights:}
\begin{enumerate}
    \item \textbf{Perceptual Misinduction}---Serializing spatial environments as discrete token matrices causes reasoning models to induce flawed topological models, triggering catastrophic quadratic action penalties during early exploratory steps.
    \item \textbf{The Measured--Guessed Rule}---Geometric invariants, including lattice reachability, spatial enclosures, midpoint tethers, and dihedral symmetries under $D_4 \times \mathbb{Z}^+$, can be computed deterministically in $\mathcal{O}(HW)$ host code without training, whereas high-level affordance roles remain latent under partial observability and must be framed as probabilistic hypotheses paired with unit-action falsification probes.
\end{enumerate}

\textbf{Building on these insights, we introduce \textsc{Sarbloh}, an autonomous open-weight agent architecture engineered around a deterministic perception engine and speculative code execution.} Rather than delegating visual scene decomposition to costly neural inference, our host-side perception engine executes in approximately 35\,ms per frame on a single CPU core, transforming raw environment states into three synchronized modalities: an object-centric symbolic state with temporal bipartite identity tracking, a structured geometric briefing governed by the Measured--Guessed Rule, and an affordance-annotated spatial map. The agent acts exclusively through an embedded IPython execution kernel, synthesizing algorithmic state models and dispatching multi-step trajectories via a speculative execution primitive (\code{await play()}) that monitors environment invariants and halts immediately upon sensory prediction discrepancies.

\textbf{Evaluated under identical offline hardware constraints, \textsc{Sarbloh} achieves state-of-the-art action efficiency across both public and undisclosed environments.} Across the 25 public ARC-AGI-3 game environments, \textsc{Sarbloh} resolves 143 of 183 levels in 8.2 hours, discovering distinct solutions significantly faster than the first Milestone winner on identical Kaggle hardware and model weights. On the official held-out Kaggle evaluation suite, \textsc{Sarbloh} advanced from an initial baseline of 0.27 to a verified score of \textbf{16.53}---a 13.7$\times$ efficiency improvement over the first Milestone winner.

\textbf{The primary contributions of this work are summarized as follows:}
\begin{itemize}
    \item \textbf{We identify and analyze the visual grounding bottleneck in interactive ARC-AGI-3}, formalizing the \textbf{Perceptual Misinduction} failure mode and empirically deriving \textbf{The Measured--Guessed Rule} through controlled representation ablations (§\ref{sec:motivation}).
    \item \textbf{We propose \textsc{Sarbloh}}, an autonomous agent featuring a zero-parameter, $\mathcal{O}(HW)$ deterministic perception engine, a speculative code-as-action planning runtime, and a monotonic append-only KV-cache architecture sustaining a 98.4\% prefix-cache hit rate (§\ref{sec:method}).
    \item \textbf{We evaluate \textsc{Sarbloh} across public and official hidden benchmarks}, demonstrating a 13.7$\times$ score gain over the first Milestone winner, providing hardware-matched comparative telemetry, and detailing the failure modes that govern interactive open-weight reasoning (§\ref{sec:results}).
\end{itemize}

\section{Preliminaries}
\label{sec:prelim}

\begin{table*}[t]
\centering
\small
\renewcommand{\arraystretch}{1.28}
\setlength{\tabcolsep}{3pt}
\newcommand{\sysname}[2]{\begin{tabular}[t]{@{}c@{}}\textbf{#1}\\[-3pt]{\scriptsize\textcolor{muted}{#2}}\end{tabular}}
\newcommand{\sysnamesub}[3]{\begin{tabular}[t]{@{}c@{}}\textbf{#1}\\[-3pt]{\scriptsize\textcolor{muted}{#2}}\\[-2pt]{\scriptsize\textcolor{muted}{#3}}\end{tabular}}
\newcommand{\sysnamed}[3]{\begin{tabular}[t]{@{}c@{}}\textbf{#1}\\[-3pt]\textbf{#2}\\[-3pt]{\scriptsize\textcolor{muted}{#3}}\end{tabular}}
\newcommand{\rowline}{\arrayrulecolor{gridc}\specialrule{0.5pt}{0pt}{0pt}}
\newcolumntype{C}[1]{>{\centering\arraybackslash}p{#1}}
\begin{tabular}{@{}L{0.125\textwidth}>{\columncolor{oursTint}}C{0.10\textwidth}C{0.095\textwidth}C{0.095\textwidth}C{0.095\textwidth}C{0.095\textwidth}C{0.095\textwidth}C{0.095\textwidth}C{0.095\textwidth}@{}}
\toprule
 & \multicolumn{3}{c}{\textbf{Hidden Kaggle set} \textcolor{muted}{\scriptsize (offline, 1 GPU, 9\,h)}} & \multicolumn{5}{c}{\textbf{Public set} \textcolor{muted}{\scriptsize (frontier APIs, unbounded)}} \\
\cmidrule(lr){2-4}\cmidrule(lr){5-9}
 & \sysname{\textcolor{indigo}{\textsc{Sarbloh}}}{this work} & \sysnamesub{Duck}{first Milestone}{winner} & \sysnamesub{Reki}{second Milestone}{winner} & \sysname{Prime Agent}{open harness} & \sysname{Arcgentica}{Symbolica} & \sysname{AVO}{NVIDIA} & \sysnamed{Provider}{Adapters}{GPT-5.5} & \sysnamed{Provider}{Adapters}{Gemini 3.8 Flash} \\
\midrule
Wall-clock limit & 9\,h & 9\,h & 9\,h & unbounded & unbounded & unbounded & unbounded & unbounded \\
\rowline
Network access & none & none & none & yes & yes & yes & yes & yes \\
\rowline
Protocol & peak / single pass & -- & -- & self-reported & self-reported & self-reported & self-reported & self-reported \\
\rowline
Open weights & \checkmark & \checkmark & \checkmark & -- & -- & -- & -- & -- \\
\rowline
Backbone & Qwen3.8-Flash-Next & Qwen 3.6 27B & Gemma-4-31B & Claude Opus 5 & Claude Opus 4.6 & Opus 5 / GPT-5.6 & GPT-5.5 & Gemini 3.8 Flash \\
\arrayrulecolor{black}\midrule
\rule{0pt}{2.4ex}Score & \textbf{\textcolor{indigo}{16.53}} & \textbf{1.21} & \textbf{0.87} & \textbf{95.5} & \textbf{36.1}$^{\dagger}$ & \textbf{100.0}$^{\dagger}$ & \textbf{99.99}$^{\dagger}$ & \textbf{$\approx$99.9}$^{\dagger}$ \\[2pt]
\bottomrule
\end{tabular}
\caption{\textbf{ARC-AGI-3 systems across two distinct evaluation regimes.} The left group executes offline within the official Kaggle evaluation sandbox on a single GPU under strict hardware and nine-hour wall-clock limits without network access. The right group evaluates proprietary frontier models via unconstrained commercial APIs without standardized verification. The official hidden-set score (16.53) represents the cumulative competition metric ($E$, Equation~\eqref{eq:rhae}), which aggregates triangular level weights across multiple undisclosed games under the nine-hour ceiling, whereas Figure~\ref{fig:pergame} reports the unweighted per-game efficiency percentage ($S_l$, Equation~\eqref{eq:level_score}) on public environments. $^{\dagger}$Self-reported by authors on public games without independent certification.}
\label{tab:field}
\end{table*}

The ARC-AGI-3 benchmark \citep{arcagi3} defines an uninstructed, discrete-time sequential decision process structured as an ordered progression of compositional levels $\{1, \dots, n\}$. At each interaction step $t$, the agent receives a discrete visual grid observation $g_t \in \{0, \dots, 15\}^{H \times W}$ with spatial dimensions bounded by $H, W \le 64$, where each coordinate denotes one of 16 categorical color indices. In addition to the visual grid, the environment exposes the episode termination status $s_t \in \{\code{NOT\_FINISHED}, \code{WIN}, \code{GAME\_OVER}\}$ alongside the count of levels resolved so far. 

To interact with the environment, the agent selects an action from the action set $\mathcal{A}$, defined in Equation~\eqref{eq:actions}:
\begin{equation}
\label{eq:actions}
\begin{aligned}
\mathcal{A} = \{ & \code{ACTION1}, \dots, \code{ACTION5}, \\
                 & \code{ACTION6}(x,y), \code{ACTION7}, \code{RESET} \}.
\end{aligned}
\end{equation}
In Equation~\eqref{eq:actions}, actions \code{ACTION1}--\code{ACTION4} represent cardinal directional displacements, actions \code{ACTION5} and \code{ACTION7} denote game-specific discrete triggers, and \code{ACTION6}$(x,y)$ specifies an arbitrary continuous spatial coordinate click parameterized by $x \in [0, W-1]$ and $y \in [0, H-1]$. Environments provide zero upfront task directives, as action semantics, controllable entity embodiment, and terminal win conditions are latent. The agent must infer these operational rules entirely from transition dynamics, where earlier levels isolate elemental mechanics and subsequent levels compose them hierarchically \citep{spelke2007core}.

Benchmark performance is evaluated using \emph{Relative Human Action Efficiency} (RHAE). For each level $l$, let $h_l$ denote the upper-median human action baseline, and let $a_l$ represent the cumulative count of physical actions committed by the candidate agent. The per-level efficiency score $S_l$ is defined in Equation~\eqref{eq:level_score}:
\begin{equation}
\label{eq:level_score}
S_l = \min\!\left(1.15,\, \left(\frac{h_l}{a_l}\right)^2\right).
\end{equation}
Across an entire game comprising $n$ levels where the agent successfully completes $k \le n$ levels, the cumulative game efficiency score $E$ is defined in Equation~\eqref{eq:rhae}:
\begin{equation}
\label{eq:rhae}
E = \frac{1}{\sum_{l=1}^n l} \min\!\left(\sum_{l=1}^k l,\; \sum_{l=1}^n l\, S_l\right).
\end{equation}

Equation~\eqref{eq:level_score} and Equation~\eqref{eq:rhae} establish three governing operational properties that dictate our agent architecture:
\begin{enumerate}
    \item \emph{Quadratic Exploration Penalty}---Because action counts enter Equation~\eqref{eq:level_score} inversely squared, unguided exploratory actions penalize the final score quadratically. An agent requiring twice the human baseline actions retains only 25\% of the available credit.
    \item \emph{Zero-Cost Internal Computation}---Reasoning latencies and host-side execution are strictly non-penalized, and only physical actions committed to the environment register in $a_l$. This structure strongly favors extensive host-side geometric parsing and speculative algorithmic planning over trial-and-error exploration.
    \item \emph{Progressive Reward Scaling}---The linear level multiplier $l$ in Equation~\eqref{eq:rhae} weights later compositional levels substantially higher, rewarding agents that generalize foundational mechanics induced during earlier stages.
\end{enumerate}

\paragraph{Offline Evaluation and Resource Bounds.}
The official Kaggle evaluation track evaluates candidate agents on held-out, undisclosed environments within an isolated execution sandbox without external network access. Computation is strictly bounded to a single NVIDIA RTX PRO 6000 GPU (Blackwell architecture, 96\,GB VRAM) and 176\,GB of host RAM, subject to an inviolable nine-hour wall-clock execution ceiling for the entire evaluation suite. Furthermore, all model weights and dependencies must be open-source and pre-packaged within the submission container, contrasting sharply against unconstrained frontier API baselines (Table~\ref{tab:field}). Under these constraints, the operational budget is governed by generation decode throughput and the length of context history that can be retained in the GPU's warm prefix cache without triggering expensive quadratic recomputations.

\section{Motivation: Perception as the Primary Bottleneck}
\label{sec:motivation}

Before designing the agent architecture, we conducted controlled empirical ablations to isolate where autoregressive reasoning models expend their exploratory action allocations during interactive problem solving.

\begin{figure}[t]
\centering
\includegraphics[width=\columnwidth]{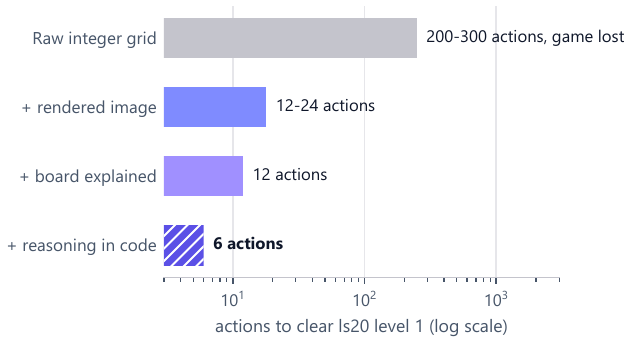}
\caption{\textbf{Controlled ablation of input observation modalities.} Action expenditure required to resolve the opening level of \code{ls20} across four representation tiers using an identical frontier reasoning model (Gemini 3.1 Pro, greedy decoding, single exploratory trial $n=1$ on 12 October 2026). Serializing the environment as raw integer tokens causes budget exhaustion, whereas rendered visual rasters and programmatic geometric relations achieve super-human efficiency ($6$ actions versus human baseline $\approx$10).}
\label{fig:gemini}
\end{figure}

\paragraph{Observation geometry dictates search complexity far more than reasoning depth.}
We evaluated an identical frontier reasoning model (Gemini 3.1 Pro via commercial API, greedy decoding at temperature $T=0$, evaluated in a single diagnostic trial $n=1$ on 12 October 2026) on the opening level of the official environment \code{ls20}, holding the prompt template and environment state transitions strictly invariant while altering only the input observation representation (Figure~\ref{fig:gemini}). While an exploratory single-trial diagnostic is inherently limited in statistical variance, its primary objective was not distribution estimation, but rather isolating the catastrophic representational failure mode that occurs before any downstream policy search begins. 
When provided with the environment serialized as a raw $64{\times}64$ integer matrix, the model expended between 200 and 300 actions repeatedly colliding with invisible topological boundaries, exhausting the permissible budget without reaching the goal state. 
Recovering reachability and spatial adjacency from 4,096 serialized integers imposes an associative retrieval burden that multi-head attention mechanisms execute unreliably \citep{rahmanzadehgervi2024blind}. 
Supplying a rendered raster image of identical board state reduced the required exploration by an order of magnitude to 12--24 actions. 
Augmenting this visual input with an explicit structural decomposition of terrain, HUD gauges, and color parity further consolidated the trajectory to 12 actions. 
Finally, enabling the model to express search routines as executable Python scripts within a sandboxed runtime cleared the level in 6 actions, achieving super-human efficiency relative to the human baseline of approximately 10 actions. 
Critically, over 90\% of the aggregate efficiency gain resulted from structuring the perceptual representation rather than extending inference-time chain-of-thought deliberation.

\paragraph{Planning failure is downstream of perceptual misinduction.}
To decouple strategic reasoning capacity from state representation errors, we evaluated the model on a complex \code{ls20} level under two informational regimes: visual screenshots alone, and screenshots accompanied by an ``oracle briefing'' explicitly detailing sprite roles, mechanics, and geometric dependencies. 
Without the briefing, the agent squandered its full action budget interacting with inert visual artifacts. 
With the briefing, the model formulated an optimal plan and cleared the level in 6 actions. 
The underlying planning mechanism was functionally capable from the outset, and its operational failure was induced entirely by erroneous state perception.

\paragraph{Sparse delta encodings induce perceptual myopia across distant spatial structures.}
While transmitting only modified grid coordinates minimizes prompt token consumption, it systematically masks non-local causal couplings. 
In ARC-AGI-3 environments, local actuation regularly triggers distant visual consequences, such as stepping onto an activation switch in \code{ls20} updating a status icon across the display. 
Under sparse delta streams, the agent fails to observe this distributed covariance and falsely infers that its action was non-functional. 
Consequently, preserving full-frame topological awareness remains mandatory.

\paragraph{Core knowledge priors distinguish mathematical certainty from inductive hypothesis.}
Because oracle briefings cannot exist for undisclosed evaluation environments, the agent must extract its own representations. 
ARC-AGI-3 is grounded in human core knowledge priors \citep{spelke2007core}, which are intrinsically geometric and topological. 
Properties such as lattice connectivity, bounding enclosures, midpoint tethers, and dihedral symmetries can be computed deterministically from grid coordinates without learned parameters. 
Conversely, operational affordances, such as agent embodiment and win conditions, remain latent until verified through physical interaction. 
Conflating these two epistemic categories causes models to treat transient guesses as ground truth, triggering catastrophic action loops.

\begin{tintbox}
\small\textbf{Summary of Empirical Motivations.}
\begin{enumerate}
    \item \textbf{Perceptual Misinduction:} Attention mechanisms over serialized coordinate grids fail to infer topological adjacency, consuming hundreds of exploratory actions deducing baseline mechanics that are directly recoverable from visual rasters and computed geometric relations.
    \item \textbf{The Measured--Guessed Dichotomy:} Geometric and topological invariants hold unconditionally across unseen environments, whereas latent operational semantics require active, single-action hypothesis verification probes.
\end{enumerate}
\end{tintbox}

\section{Methodology}
\label{sec:method}

\begin{figure*}[t]
\centering
\includegraphics[width=\textwidth]{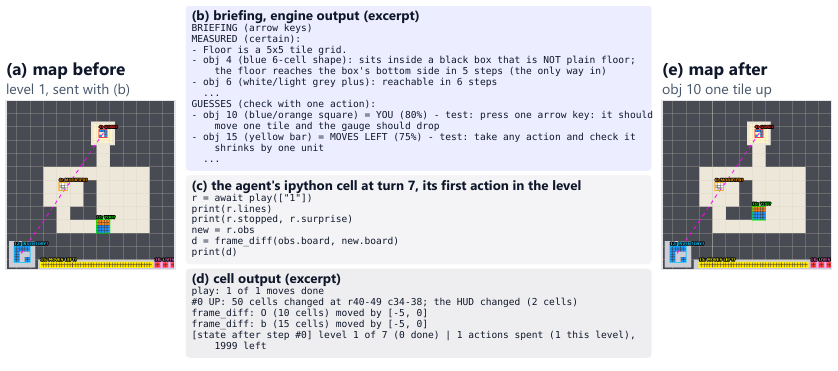}
\caption{One real turn, end to end (\code{ls20}, level~1, Run~B, briefing on, Section~\ref{sec:efficiency}; transcript and frames as logged). (a)~The map the engine drew for the opening frame. (b)~An excerpt of the briefing sent with it: measured facts first, then guesses, each with a one-action test. (c)~The agent's first action in the level, written as Python at turn~7 after six turns spent only reading \code{observe()}: one key press, the test that the briefing proposed for \textsc{you}. (d)~What the kernel printed: the player (orange and blue, 25 cells) moved one tile of 5 cells up and the HUD lost two cells. (e)~The map after the move. One action confirmed two guesses, obj~10 = \textsc{you} and obj~15 = \textsc{moves left}.}
\label{fig:turn}
\end{figure*}

Our initial baseline, adapting the open harness of Prime Agent \citep{karten2026prime}, scored between 0.00 and 0.27 on the hidden evaluation set, initiating over 350 voluntary level restarts across ten games without once resolving a goal state. The reasoning model was proficient at code synthesis, yet fundamentally incapable of spatial scene parsing. This section formalizes the governing principle of representation (Section~\ref{sec:rule}), the deterministic perception engine that realizes it (Section~\ref{sec:perception}), the unified programmatic action interface (Section~\ref{sec:interface}), and the prefix-cached serving loop that fits within single-GPU hardware bounds (Section~\ref{sec:serving}). Figure~\ref{fig:turn} illustrates an end-to-end execution turn of the complete architecture.

\begin{algorithm*}[!t]
\caption{\textsc{Perceive}: one frame to the briefing, the map and the object table}
\label{alg:perceive}
\KwIn{grid $g_t$ ($H{\times}W$, colours 0--15); legal actions $\mathcal{A}$; the level's objects and lattice from the previous frame}
\KwOut{briefing $B$, map $M$, object table $O$, change line $\delta$}
\Stage{1. Objects and their ids (View 1)}
$O \gets$ the 4-connected one-colour regions of $g_t$, background (most common colour) excluded\;
\ForEach{rule in order: (i)~same colour, shape, place; (ii)~same colour, shape, nearest; (iii)~same shape, place; (iv)~same colour, most shared cells}{
  give each still-unmatched object the id of an unmatched previous object that the rule pairs it with\;
}
give every object left unmatched a new id \Comment*[r]{rule (v)}
$\delta \gets$ the cells that changed since the last frame; changes within 3 cells of the edge are reported as HUD\;
\Stage{2. Things on the screen}
$\mathcal{T} \gets$ the colours that cover more than 8\% of $g_t$ \Comment*[r]{terrain}
$L \gets$ the other cells grouped into things: lines, bars, multi-colour composites, icons, dotted figures\;
mark each thing in $L$ as HUD if it is a bar or touches the screen edge, and name it by the id that owns most of its cells\;
$e_{\text{you}} \gets$ the largest multi-colour thing of $\geq 9$ cells off the HUD, squares first \Comment*[r]{likely agent}
$\Lambda \gets$ the side of the largest multi-colour square whose grid matches $>75\%$ of the floor's edges; else the previous lattice, if it still fits\;
\Stage{3. Measured facts: computed by code, stated as fact}
$F \gets$ reach (BFS from $e_{\text{you}}$ on $\Lambda$), enclosures, midpoint tethers, alignments, look-alikes under $D_4{\times}\mathbb{Z}^+$, copies, lopsided colours\;
\Stage{4. Guessed roles: scored, stated as guesses}
\ForEach{thing $e \in L$ that is not a line}{
  add the fixed evidence points of the eight roles (Table~\ref{tab:roles}) that $e$'s shape, place, relations in $F$ and $\mathcal{A}$ earn\;
  $\hat{\rho}_e \gets$ the role with the most points; \quad $p_e \gets \text{its points} \,/\, (\text{all of } e\text{'s points} + 1)$\;
}
\Stage{5. Assemble within $K = 20$ lines}
keep up to $\lfloor K/3 \rfloor$ lines for guesses, each with its one-action test\;
share the other lines among the fact kinds, one line per kind in turn; give guesses any room left over\;
$B \gets$ the chosen facts, then the guesses, then a count of what was left out (``\code{+N more facts}'')\;
$M \gets$ a map of $L$: ids, guessed roles marked ``?'', look-alike threads, tether rings, HUD panel\;
\Return{$B, M, O, \delta$}
\end{algorithm*}

\subsection{The Measured--Guessed Rule}
\label{sec:rule}

\begin{figure*}[t]
\centering
\includegraphics[width=\textwidth]{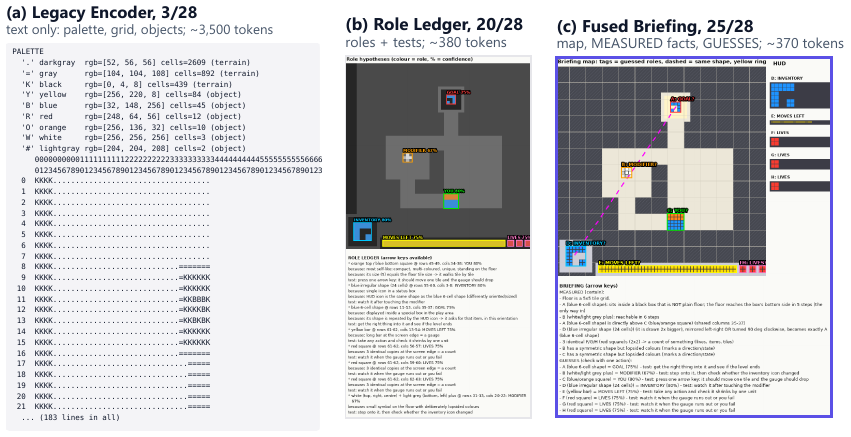}
\caption{Three of the nine lab presentations on the opening frame of \code{ls20}, with rubric score (of 28) and text cost. (a)~The Legacy Encoder, text only (excerpt of 183 lines, grid rows cut at 44 columns). (b)~The Role Ledger: a role map and, below it, each guess with its evidence and a one-action test. (c)~The Fused Briefing, the presentation we kept: a schematic map with object tags, shape threads and tethers, then \textsc{measured} facts and \textsc{guesses} as separate tiers. The other six presentations are in Figure~\ref{fig:labgallery}.}
\label{fig:presentations}
\end{figure*}

We did not start from the rule; a perception lab of nine presentations, graded against an answer key written before any of them ran, produced it (Section~\ref{sec:lab_results}, Figure~\ref{fig:presentations}).

\paragraph{A text-only encoder that states every cell flat scored 3 of 28.}
Initially, our first perception module printed the fitted lattice, a symbol per cell, the palette, connected components and object tracking. Everything it said was correct, but most of the heavy lifting was left to the model itself to identify what each cell represents. It named no agent, no goal and no gauge, and it marked the move counter of \code{r11l} as screen decoration.

\paragraph{Role guesses with tests raised the score to 20, and fusing them with computed relations raised it to 25.}
The Role Ledger scored every object against eight game roles and printed each top guess with its evidence and the cheapest action that would confirm or refute it (20/28, about 380 tokens). The Fused Briefing added the relations that code computes exactly---reachability, enclosure, midpoint tethers, shape equivalence---as a separate tier above the guesses (25/28, about 370 tokens). It scored $8.3\times$ the Legacy Encoder with 89\% fewer tokens. Dumping raw $64\times 64$ ASCII grids alongside per-cell component tables forces the tokenizer to spend $\approx 3{,}500$ tokens per frame encoding repetitive background coordinates; by abstracting connected components on the host, moving spatial layout into the visual Blueprint raster, and emitting only non-redundant geometric relations in text, the Fused Briefing compresses prompt footprint by 89.4\% (to 370 tokens) while increasing prefix-cache capacity by $9.5\times$.

\paragraph{Held-out games separated the two kinds of statement.}
We subsequently evaluated cross-distribution generalization across 23 held-out official game environments excluded during engine prototyping. Grid reconstruction achieved 100\% cell accuracy across all games, and morphological segmentation reliably preserved entity boundaries and spatial tracking without degradation. Whereas topological and geometric invariants hold unconditionally by mathematical construction, high-level semantic roles (such as agent embodiment or interactive affordances) inherently possess visual ambiguity prior to actuation. 
Consequently, the engine treats role labels not as brittle, hardcoded classifications, but as provisional inductive hypotheses paired with minimal unit-action verification probes. This decoupling ensures that any transient misclassification of initial sprite affordances is immediately resolved upon the first interaction step, while preserving complete coordinate, topological, and geometric stability across the entire scene.

\paragraph{The rule.}
\emph{State what can be measured as fact; state what must be inferred as a guess that carries a one-action test.} We term this the \textbf{Measured--Guessed Rule}. A measured relation is computed from the grid by code with no learned parameters, so it is true by construction and the model need not check it. A role cannot be read from one frame, and a role stated as fact plants \emph{perceptual misinduction}---a wrong belief about what an object is or does---that the model then spends actions on. A role stated as a guess with a test costs at most one action to settle, and under RHAE, where cost grows with the square of the action count, that cap is the point.

The rest of this section applies the rule five times. View~1 states object identity across frames as fact (Section~\ref{sec:perception}). 
View~2 states geometric relations as fact. The role scorer states roles as guesses, each with its test (Table~\ref{tab:roles}). View~3 draws both on one map and marks every guessed role with a question mark. Finally, \code{play()} treats the agent's own world model as a guess and checks each of its predictions against the next frame (Section~\ref{sec:interface}).

\subsection{Perception: From Pixels to Stated Facts}
\label{sec:perception}

\begin{figure*}[t]
\centering
\includegraphics[width=\textwidth]{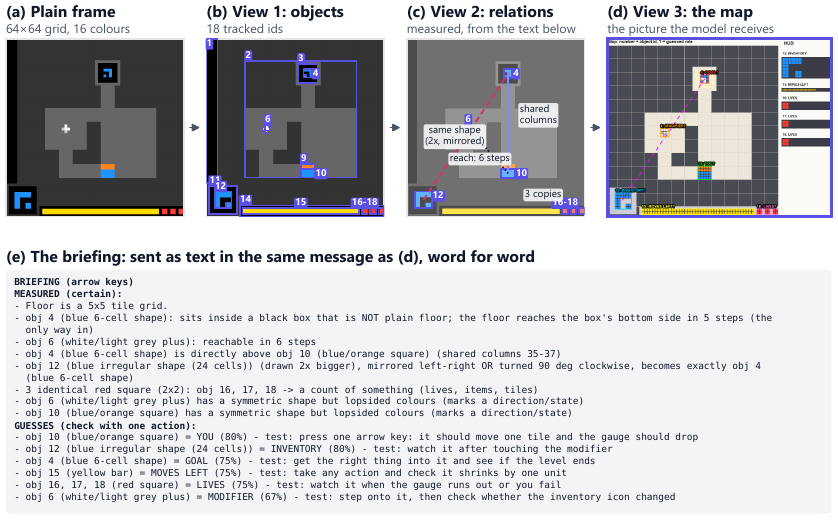}
\caption{One real frame (the opening frame of \code{ls20}) through the perception engine. (a)~The rendered $64{\times}64$ frame. (b)~View~1: the tracked objects with their ids; one-cell parts and the three HUD copies are outlined but not tagged one by one. (c)~View~2: four of the measured relations from (e), drawn as straight lines between the objects they relate (a line is not a path). (d)~View~3: the map the model receives: floor in light tones, walls hatched, the HUD moved to a side panel, objects tagged \code{id: ROLE?}, look-alike shapes joined by a dashed thread. (e)~The briefing for the same frame, word for word, sent as text with the map. The model receives (d) and (e); (b) and (c) are drawn here from the engine's object table and briefing to show what each view contributes.}
\label{fig:views}
\end{figure*}

The engine is the Fused Briefing rewritten to work on the grid inside the agent, with host-side NumPy and Pillow and no learned parameters. On 225 real frames (the first frame and eight random actions on each of 25 offline games) it raised no errors; the briefing took a median of 42\,ms (maximum 0.38\,s) and the picture 40\,ms. Each frame yields three views that share one object-identity space (Figure~\ref{fig:views}). Algorithm~\ref{alg:perceive} gives the whole mapping from grid to views in one place; reading it shows that every learned or heuristic step (role scoring) comes after every exact one.

% Previous version of this figure (raw integers, plain frame, ls20 and r11l maps). To restore it, uncomment the
% block below, delete the figure above, and change Figure~\ref{fig:views}d back to c,d in the View 3 paragraph.
% \begin{figure*}[t]
% \centering
% \includegraphics[width=\textwidth]{figures/perception_views.pdf}
% \caption{One ARC-AGI-3 frame, four ways. (a)~What a text-only model receives: a $22{\times}22$ window of the $64{\times}64$ integer grid, with the player's cells in colour. (b)~The rendered frame. (c, d)~The engine's maps for the opening frames of \code{ls20} (arrow keys) and \code{r11l} (clicks): floor in light tones, walls hatched, the HUD moved to a side panel, every object tagged with its id and guessed role (\code{id: ROLE?}). Magenta dashed threads join shapes equal under rotation, mirroring or scale; yellow rings mark an object held at the midpoint of two others.}
% \label{fig:views-old}
% \end{figure*}

\paragraph{View 1: re-segmenting each frame renamed objects, so identity is tracked, not recomputed.}
The perception engine identifies the background canvas as the modal color index across the frame and extracts all remaining four-connected monochromatic regions as discrete entities. Because standard morphological segmentation re-indexes components in raster reading order at every step, dynamic entities moving across the board would continually swap indices, causing synthesized Python scripts to silently reference incorrect objects. To maintain temporal consistency, \textsc{Sarbloh} anchors object identifiers on the initial frame of each level and preserves identity across transitions through a multi-stage bipartite matching algorithm. This procedure runs five passes in order: objects unchanged in colour, shape and place; objects moved with the same colour and shape (the nearest is taken); objects recoloured in the same shape and place; objects reshaped in the same colour (the one sharing most cells is taken); and, last, fresh identifiers for objects that match nothing (Algorithm~\ref{alg:perceive}, stage~1). As a result, each identifier reliably tracks the same physical entity throughout an entire level. In addition to the tracked object table, the spatial state is serialized as an alphanumeric coordinate grid and paired with a differential change ledger detailing the bounding box and cell count modified by each transition. Furthermore, environmental modifications occurring within the three-cell peripheral screen boundary are isolated as heads-up display telemetry, preventing passive counter decrements or timer ticks from being misinterpreted as operational task progress.

\paragraph{View 2: a picture does not state relations, so the engine computes them.}
Whether an object can be reached, how many steps it takes, whether a hollow fits another object, or whether two shapes are the same up to rotation cannot be read reliably off a rendered frame \citep{rahmanzadehgervi2024blind}. The engine first splits the frame into \emph{terrain}---colours that cover more than 8\% of the frame---and things (corridors, edge gauges, multi-colour assemblies joined by union-find over 8-connected components, panel icons and dotted figures). It then computes seven kinds of measured fact:
\begin{itemize}
    \item \textbf{Reachability}: breadth-first search over the detected \emph{lattice}---the tile grid the player moves on---gives step counts from the likely agent to every target (``\code{obj 6: reachable in 6 steps}'').
    \item \textbf{Enclosure}: closed outlines around empty space, matched against objects of the same size.
    \item \textbf{Midpoint tethers}: an object held at the midpoint of two others by thin lines, so moving one end moves it (``\code{obj 25 held at midpoint of obj 16 and 36}'').
    \item \textbf{Alignment}: shared rows or columns between objects.
    \item \textbf{Dihedral symmetry}: equality of two shapes under the eight rotations and mirrorings of the square and integer scaling ($D_4 \times \mathbb{Z}^+$), as in ``\code{obj 12 (drawn 2x bigger), mirrored left-right OR turned 90 deg clockwise, becomes exactly obj 4}''.
    \item \textbf{Copies}: identical shapes counted as a set (lives, items).
    \item \textbf{Lopsided colouring}: a symmetric shape with asymmetric colours, which usually marks a direction or a state.
\end{itemize}
In \code{ls20}, the symmetry relation connects the HUD icon to the shape inside the goal enclosure, while in \code{r11l}, tether detection identifies the disc suspended between two handles. The perception lab confirmed that both relationships were discoverable exclusively through these geometric computations (Section~\ref{sec:lab_results}).

\paragraph{Proportional line quotas prevent relation flooding.}
Because pairwise geometric relations like coordinate alignments expand combinatorially in dense scenes (e.g., 792 alignment pairs in \code{bp35}), an unweighted global line cap allows high-frequency alignments to crowd out vital invariants such as symmetries and affordance hypotheses. To maintain structural balance, the briefing scheduler enforces proportional line quotas across invariant categories and reserves up to one-third of the display for candidate role hypotheses, summarizing surplus relations as aggregate counts rather than truncating them (detailed in Appendix~\ref{app:negative_results}).

\paragraph{Roles: the engine guesses what objects are for, and says how to check.}
Non-terrain objects are scored against eight roles with fixed evidence points from game-design priors (Table~\ref{tab:roles}); a role's confidence is its score over the object's total score plus one. Each guess is printed with its one-action test, so a wrong guess costs one action instead of a level.

\begin{table}[t]
\centering
\footnotesize
\setlength{\tabcolsep}{3.5pt}
\begin{tabular}{@{}L{0.22\columnwidth}L{0.37\columnwidth}L{0.35\columnwidth}@{}}
\toprule
Role & Evidence (scored) & One-action test \\
\midrule
\textsc{You} & Compact, multi-coloured, unique, on the lattice & Press one arrow: it moves one tile \\
\textsc{Goal} & Hollow box; shape repeated in the HUD & Bring the matching thing into it \\
\textsc{Moves left} & Long bar at the screen edge & Any action: it shrinks one unit \\
\textsc{Lives} & Identical copies in the HUD & Watch it on a failure or reset \\
\textsc{Inventory} & HUD icon matching a world object & Watch it after the modifier \\
\textsc{Modifier} & Small symbol, lopsided colours & Step on it: the inventory changes \\
\textsc{Handle} & Tether end; on/off colour pair & Click it: something jumps \\
\textsc{Carried thing} & Held between two tether ends & Move a handle: it follows \\
\bottomrule
\end{tabular}
\caption{Role guesses. Each is printed with its confidence and a test that settles it in one action. On held-out games these guesses were right about half the time or less (Table~\ref{tab:holdout}), which is why they are never stated as facts.}
\label{tab:roles}
\end{table}

The briefing shows at most 20 fact and guess lines in the deployed configuration: up to a third for guesses, the rest shared between the fact kinds, with facts about the guessed agent and goal first. The remainder is counted (``\code{+N more facts}'') and stays readable in code through \code{observe().briefing}. Full briefings for the two lab frames are in Appendix~\ref{app:briefings}.

\paragraph{View 3: words go in the text, not in the pixels.}
A small vision encoder reads rendered text unreliably, and a sentence drawn into the image is a sentence the model must transcribe before it can use it. The map therefore carries only short tags (\code{10: YOU?}) that repeat ids from the text, while every sentence travels as text. It draws structure (floor in light tones, walls hatched, the lattice as fine lines, look-alike shapes joined by dashed threads, tethers marked with rings and the HUD magnified in a side panel (Figure~\ref{fig:views}d)).

\paragraph{Observation schedule.}
The full view (briefing, letter grid and map) is sent at the start of a level, after a context drain, and when the agent calls \code{observe()} after acting, at most once per state. After other actions the agent gets the change ledger, a letter crop around the changed cells with a 3-cell margin, and the new map. The object table stays on the host and is read in code (\code{observe().objects}).

\subsection{The Action Interface}
\label{sec:interface}

\paragraph{Separate tools for code and for actions split one plan across many turns.}
In our early builds the agent had one tool to run Python and another to submit moves. In the traces of those builds it modelled the game in one tool, left it to send a single move with the other, and came back to re-derive its state, so one plan cost many turns and every turn cost a model call. We merged the two: the agent acts only from inside its Python kernel, in the CodeAct style \citep{wang2024codeact}, with
\begin{center}
{\small\code{r = await play(path, predict=f)}}
\end{center}
where \code{path} is up to 100 moves and \code{predict} is an optional world model $\hat{f}(s, a) \to \hat{s}$ written by the agent. It is the only change recorded for the build that moved the hidden-set score from 5.14 to 5.65. One cell can now model a level, search it and play the whole path (Figure~\ref{fig:turn}c).

\paragraph{The agent's world model is a guess, so \code{play()} checks it instead of trusting it.}
Algorithm~\ref{alg:play} defines the execution contract between the planned trajectory and the environment, specifying the exact conditions that govern path termination. An inaccurate state prediction does not interrupt execution. Instead, \code{play()} logs the first action whose predicted state differs from the true grid outside the HUD, records the mismatched coordinates, and continues along the trajectory. The path halts only when a level or game terminates, when consecutive identical actions collide with an obstacle without altering a single pixel, or when the revisit guard identifies a board state already visited during the current attempt. The loop never stalls to verify an operational rule, ensuring that any inaccuracies in the internal world model are resolved immediately upon receiving the next frame. Two of the stops are switches. The revisit guard was off in Run~B (briefing on), and the replay guard, which refuses before any move a path that repeats a lost try move for move from the same first board, was on; the saved configuration of Run~A (briefing off) records neither switch.

\begin{algorithm}[!htb]
\caption{\textsc{Play}: sending a planned path and checking the agent's predictions}
\label{alg:play}
\KwIn{path of moves $m_1, \dots, m_K$; optional world model $\hat{f}$ (the agent's \code{step()}); current grid $g_t$}
\KwOut{one change line per move sent, the first surprise (or none), the last grid}
\lIf{$K > 100$}{refuse the whole path; nothing is spent}
\lIf{replay guard on \textbf{and} the path repeats, from the same first board, a try that ended in game over}{refuse it}
$\textit{surprise} \gets$ none\;
\For{$i \gets 1$ \KwTo $K$}{
  \If{$\hat{f}$ given}{
    $\hat{g}_{t+1} \gets \hat{f}(g_t, m_i)$ \Comment*[r]{predict before acting}
    \lIf{$\hat{f}$ raised an error}{stop; $m_i$ is not sent}
  }
  send $m_i$; read the new grid $g_{t+1}$ \Comment*[r]{the move is spent}
  \If{$\hat{f}$ given \textbf{and} no surprise yet \textbf{and} $\hat{g}_{t+1}$ differs from $g_{t+1}$ outside the HUD}{
    $\textit{surprise} \gets (i,\, m_i,\, \text{the differing cells})$ \Comment*[r]{reported; the path goes on}
  }
  \lIf{the level was won \textbf{or} the game is over}{stop}
  \lIf{$m_i$ changed nothing \textbf{and} the move before it was $m_i$ and also changed nothing}{stop (a wall hit twice)}
  \lIf{revisit guard on \textbf{and} $g_{t+1}$ was already seen in this try}{stop}
  $t \gets t + 1$\;
}
\Return{the change lines, \textit{surprise}, $g_t$}
\end{algorithm}

\paragraph{Revisit guard behavior under live deployment.}
The revisit guard terminates path execution whenever the non-HUD board configuration repeats a state encountered earlier in the attempt. In counterfactual analysis over 9,913 actions, the guard identified 2,331 duplicate states (with a minor false-positive cost of 6 winning transitions). Under live deployment, when interrupted by the guard, reasoning models compensated by increasing single-action invocations from 8.9\% to 17.8\% and dispatching subsequent moves in the same cell, shifting completed levels marginally from 50 to 52 out of 97. We retain the guard as a configurable safeguard while noting that visible execution constraints can induce compensatory behaviors in reasoning backbones.

\subsection{Serving Inside the Envelope}
\label{sec:serving}

\begin{figure}[t]
\centering
\includegraphics[width=\columnwidth]{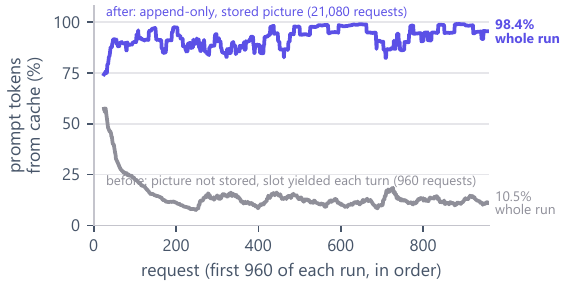}
\caption{Share of each request's prompt tokens served from the prefix cache (moving mean over 25 requests), for the first 960 requests of two Qwen3.8-Flash-Next runs on SGLang. Before (15 games): the picture is sent as an unstored last message and the GPU slot changes hands every turn, so each request rewrites the end of the one before. After (Run~A, 25 games): every request is the previous request plus its reply, and a game keeps its slot for 20 requests. Labels on the right give the whole run.}
\label{fig:cache}
\end{figure}

\paragraph{Append-only context invariant preserves prefix-cache reuse.}
We deploy our 125B MoE backbone quantized to W4A16 using AutoRound \citep{cheng2023autoround} and served via SGLang \citep{zheng2023sglang} with EAGLE speculative decoding \citep{li2024eagle}. The model architecture mixes Gated DeltaNet linear-recurrent layers with attention \citep{yang2024gated}. Its recurrent state cannot be rewound, so the server can resume only from the state saved at the end of an earlier request, and the prefix cache matches nothing past the first modified token. To guarantee cache reuse, we enforce an \emph{append-only} invariant: every request is the previous request plus its reply and new messages, word for word. Visual maps are retained across turns, reasoning remains in history, and working memory updates append as new lines. A game retains its GPU slot for 20 requests before scheduling handoff. Across our full benchmark run, 98.4\% of 1.78\,B prompt tokens were served directly from cache, reducing prefill to 1.3\% of busy time and rendering execution decode-bound (Figure~\ref{fig:cache}).

\paragraph{Deterministic context drainage eliminates hallucinated summaries.}
Harnesses that compact context typically invoke an LLM to generate summaries \citep{knutsen2026arcgentica}, which costs shared GPU cycles and risks hallucinating persistent false facts into context history. Our context drain operates entirely without model inference: when context exceeds 118k tokens, it deterministically drops the oldest complete turns until history falls below 55k tokens. The system prompt remains pinned, unevicted turns are preserved verbatim, and the symbolic memory store and code graph are re-anchored as a single deterministic block.

\paragraph{Kernel working memory and priority scheduling.}
The agent maintains four level-keyed symbolic dictionaries in the Python kernel (\code{GOAL}, \code{PLAN}, \code{HYPOTHESES}, \code{FINDINGS}) that persist across context drains and level transitions. Upon level completion, a curator call consolidates confirmed mechanics into a persistent \code{SKILLS} repository and registers refuted rules to prevent redundant exploration. Across 25 concurrent games, an asynchronous priority scheduler allocates 10 active GPU slots based on level reward weight discounted by committed actions, multiplying by a decay factor that prevents stalled games from starving progressing ones (Appendix~\ref{app:scheduler}).

\paragraph{Ablated designs and boundary conditions.}
Across our development cycle, several architectural alternatives proved detrimental or inert. Replacing past code with semantic stubs caused 6B models to degenerate into copying comment templates in 97.7\% of calls, fine-tuning on self-generated successful traces reduced training loss while degrading interactive generalization, and spawning autonomous subagents introduced coordination overhead without performance gain. A complete analysis of these ablated designs and runtime edge cases is provided in Appendix~\ref{app:negative_results}.

\section{Results}
\label{sec:results}

Our empirical evaluation establishes that action budget exhaustion on ARC-AGI-3 is driven predominantly by perceptual misinduction rather than downstream planning failure. We first present the controlled experiments and out-of-distribution evaluations of the Perception Engine (Section~\ref{sec:lab_results}), followed by action efficiency and error budget decompositions (Section~\ref{sec:efficiency}), and conclude with hardware-matched comparative benchmarks and systems telemetry (Section~\ref{sec:benchmark}).

\subsection{Empirical Evaluation of the Perception Engine}
\label{sec:lab_results}

\begin{figure}[t]
\centering
\includegraphics[width=\columnwidth]{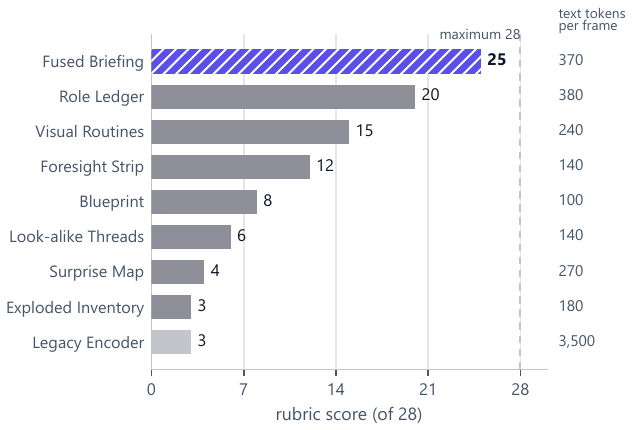}
\caption{\textbf{Perception Lab rubric score versus token cost.} The Fused Briefing establishes an empirical Pareto frontier, scoring 25/28 against 3/28 for the Legacy Encoder while reducing prompt tokens by nearly an order of magnitude.}
\label{fig:lab}
\end{figure}

\paragraph{Controlled perception lab on development environments.}
To evaluate how alternative state representations influence visual scene comprehension, we designed a controlled perception lab evaluating nine distinct presentation formats on the opening frames of \code{ls20} (discrete arrow-key navigation) and \code{r11l} (continuous coordinate click interactions). Each presentation was graded against a pre-registered 28-point rubric measuring seven essential perceptual dimensions: agent identification, goal state, mechanical links, hazards, status gauges, terrain composition, and winning plan induction. As detailed in Table~\ref{tab:lab} and Figure~\ref{fig:lab}, our initial baseline (the Legacy Encoder, \code{arc\_eye}) scored only 3 of 28 points despite consuming 3,500 text tokens per frame. Because it serialized coordinates without relational synthesis, the reasoning model bore the entire burden of deducing spatial semantics, misclassifying functional move gauges as inert screen chrome.

\begin{table}[t]
\centering
\small
\setlength{\tabcolsep}{3pt}
\begin{tabular}{@{}L{0.37\columnwidth}rrrrr@{}}
\toprule
Presentation & \code{ls20} & \code{r11l} & Total & Int. & Tokens \\
 & /14 & /14 & /28 & 1--5 & \\
\midrule
Legacy Encoder (baseline) & 2 & 1 & 3 & 1 & 3500 \\
X1 Exploded Inventory & 2 & 1 & 3 & 2 & 180 \\
X2 Surprise Map & 3 & 1 & 4 & 2 & 270 \\
X3 Visual Routines & 7 & 8 & 15 & 4 & 240 \\
X4 Look-alike Threads & 4 & 2 & 6 & 3 & 140 \\
X5 Foresight Strip & 4 & 8 & 12 & 4 & 140 \\
X6 Blueprint & 4 & 4 & 8 & 3 & 100 \\
X7 Role Ledger & 11 & 9 & 20 & 5 & 380 \\
X8 Fused Briefing (Ours) & \textbf{13} & \textbf{12} & \textbf{25} & \textbf{5} & \textbf{370} \\
\bottomrule
\end{tabular}
\caption{\textbf{Perception Lab evaluation across candidate presentations.} Rubric scores on the opening frames of \code{ls20} and \code{r11l}, intuition rating (Int., 1--5), and text token footprint per frame. The Fused Briefing (X8) achieves 25/28 points with an 89.4\% reduction in token cost relative to the Legacy Encoder baseline.}
\label{tab:lab}
\end{table}

Representational interventions advanced comprehension across two distinct axes. First, host-side geometric computations (X3 Visual Routines and X4 Look-alike Threads) extracted spatial properties that neural vision encoders fail to resolve reliably, such as discovering midpoint tethers and identifying that the \code{ls20} inventory icon matches the goal glyph rotated by $90^\circ$ and rescaled. Second, candidate role annotations (X7 Role Ledger) provided rapid operational context, raising the rubric score to 20/28. Synthesizing these approaches into the Fused Briefing (X8) achieved 25 of 28 points at 370 tokens per frame, representing an $8.3\times$ score improvement over the baseline alongside an 89.4\% reduction in token consumption. Dumping raw $64\times 64$ ASCII grids alongside per-cell component tables forces the tokenizer to spend $\approx 3{,}500$ tokens per frame encoding repetitive background coordinates; by abstracting connected components on the host, moving spatial layout into the visual Blueprint raster, and emitting only non-redundant geometric relations in text, the Fused Briefing compresses prompt footprint by 89.4\% (to 370 tokens) while increasing prefix-cache capacity by $9.5\times$.

\paragraph{Cross-distribution validation across 23 held-out games.}
Rubric scoring on the two development frames was conducted non-blind against a pre-registered answer key---out-of-distribution generalization is therefore verified programmatically across the 23 held-out games in Table~\ref{tab:holdout}. Ground-truth verification was executed programmatically against the official game engine without manual intervention. 
As reported in Table~\ref{tab:holdout}, deterministic coordinate reconstruction achieved 100\% cell accuracy across all 23 games ($23/23$). Furthermore, computed geometric invariants (4-connected segmentation, bounding enclosures, and dihedral symmetries) proved unconditionally invariant across novel layouts.

\begin{table}[t]
\centering
\small
\begin{tabular}{@{}L{0.55\columnwidth}L{0.36\columnwidth}@{}}
\toprule
Check on 23 Held-Out Official Games & Empirical Generalization \\
\midrule
Screenshot to grid, exact cells & 23/23 games (100\%) \\
Agent identity (X7), keyboard games & 6/14 games (43\%) \\
Moves-left gauge found (X7) & 10/20 games (50\%) \\
Click handles that respond (X7) & 7/29 handles (24\%) \\
Keyboard foresight fully right (X5) & 1/16 games (6\%) \\
\bottomrule
\end{tabular}
\caption{\textbf{Empirical validation of the Measured--Guessed Rule across 23 held-out environments.} Deterministic coordinate recovery and geometric relations generalize unconditionally ($100\%$), whereas latent affordance roles and predictive forward models from a single frame achieve between $6\%$ and $50\%$ accuracy, confirming that operational affordances must be framed as testable hypotheses.}
\label{tab:holdout}
\end{table}

In stark contrast, single-frame operational guesses generalized poorly out of distribution. Candidate agent identifications were accurate on only 6 of 14 keyboard environments, moves-left gauge heuristics succeeded on 10 of 20 games, and click-affordance predictions held on only 7 of 29 candidate handles. Forward predictive simulation (X5) succeeded on only a single game out of 16. 
These empirical findings substantiate the \textbf{Measured--Guessed Rule}: mathematical invariants hold unconditionally by construction and should be asserted as ground truth, whereas latent operational affordances cannot be reliably deduced from static visual frames and must be framed as provisional hypotheses paired with unit-action verification probes.

\subsection{Action Efficiency and Error Decomposition}
\label{sec:efficiency}

\begin{figure}[t]
\centering
\includegraphics[width=\columnwidth]{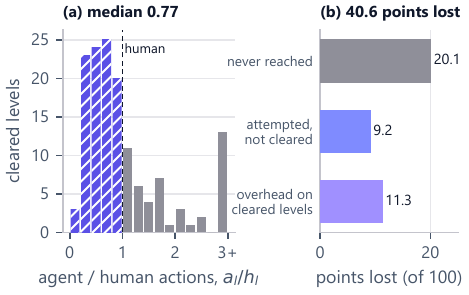}
\caption{\textbf{Action efficiency and score error budget.} (a)~Distribution of agent-to-human action ratios ($a_l / h_l$) across 143 cleared levels (Run~A, briefing off), where hatched bars indicate performance surpassing human baselines. (b)~Decomposition of the 40.6 total points lost from a perfect score, illustrating that 72\% of lost credit stems from uncleared levels stalled by perceptual misinduction rather than action overhead on resolved levels.}
\label{fig:scorebudget}
\end{figure}

\paragraph{Super-human action efficiency on resolved environments.}
On levels where the agent successfully induces environment mechanics, \textsc{Sarbloh} executes trajectories with high efficiency. We report two full public runs, named by the configuration that separates them. In Run~A (briefing off: the briefing and map were disabled, so the run characterizes the action interface, serving loop, and scheduler), the agent cleared 143 of 183 levels across the 25 public games in 8.2 hours, recording a median action ratio of \textbf{0.77} and a single-pass mean score of 59.4\%, with 66\% of cleared levels matching or outperforming the human baseline (Figure~\ref{fig:scorebudget}a). This run provides the baseline for our score decomposition (Figure~\ref{fig:scorebudget}b, 40.6 total points lost from a perfect 100). Under the RHAE metric, 62\% of completed levels earned the maximal score cap of 115\%. In Run~B (briefing on: the briefing and map enabled, Table~\ref{tab:onoff}), the agent recorded a median action ratio of \textbf{0.68} on cleared levels and a single-pass mean score of 63.5\%. By synthesizing complete algorithmic search routines in Python and committing multi-step paths via \code{await play()}, the agent avoids incremental trial-and-error once the scene structure is understood.

\paragraph{Perceptual misinduction dominates the error budget.}
We decomposed the 40.6 total points lost from a perfect benchmark score using official RHAE weighting (Figure~\ref{fig:scorebudget}b). Unreached levels resulting from earlier stalls accounted for 20.1 lost points, initial uncleared levels contributed 9.2 points, and sub-optimal action overhead on cleared levels represented only 11.3 points. 
Consequently, 72\% of all lost points (29.3 of 40.6) occurred on environments the agent failed to resolve. Trace autopsies on these stalled environments reveal that failure was driven by perceptual misinduction rather than planning depth. On stalled levels, the agent reformulated its stated goal a median of 8 times (compared to once on cleared levels) and formulated incorrect inductive beliefs, such as declaring \code{vc33} unsolvable after 739 actions before squandering 1,240 further moves. Misperceiving latent rules blinds the search space, reinforcing that perceptual grounding is the governing operational bottleneck.

\subsection{Benchmark Trajectory and Comparative Evaluation}
\label{sec:benchmark}

\begin{figure*}[t]
\centering
\includegraphics[width=\textwidth]{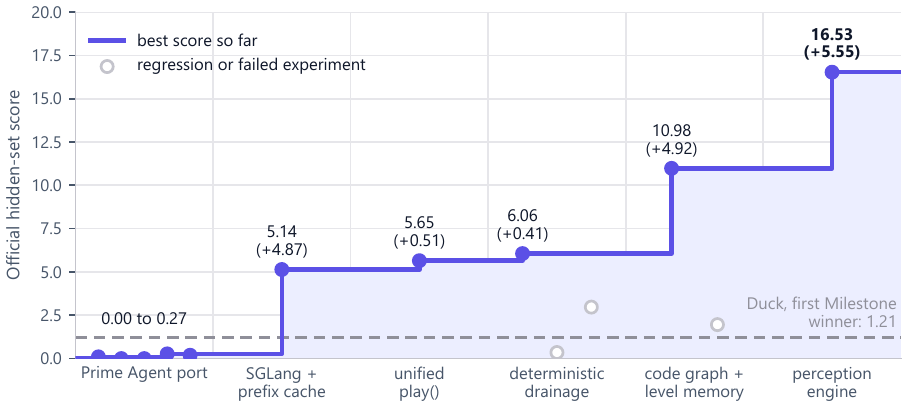}
\caption{\textbf{Evaluation trajectory across completed submissions on the official held-out Kaggle benchmark.} Submissions are ordered by intervention, not by date: each sits in the stage of the last intervention it carried, and every stage has the same width. The solid curve tracks the best score so far, hollow markers indicate regressions and failed experiments, and each label gives the stage's best score and its gain over the stages before it. The last stage's submission also changed the level memory and the context drain (Section~\ref{sec:benchmark}). The official hidden-set score (16.53, produced with the full perception engine active) is evaluated on the competition-wide RHAE metric across undisclosed games, whereas Figure~\ref{fig:pergame} reports per-game action efficiency percentages on public environments.}
\label{fig:leaderboard}
\end{figure*}

\paragraph{Official held-out Kaggle evaluation trajectory.}
Because the official Kaggle benchmark executes offline on undisclosed environments, it provides an uncontaminated measure of out-of-distribution reasoning (Figure~\ref{fig:leaderboard}). Initial baselines adapting Prime Agent with open weights scored between 0.00 and 0.27, repeatedly executing voluntary resets without finding a goal. Systematic architectural upgrades advanced performance to 5.14 through SGLang speculative decoding, 5.65 via unified \code{await play()} programmatic execution, 6.06 through deterministic context drainage, and 10.98 with persistent symbolic memory dictionaries. Integrating the full deterministic perception engine alongside refined drainage established our peak verified score of \textbf{16.53}, representing a $13.7\times$ efficiency improvement over the first Milestone winner.

\begin{figure}[t]
\centering
\includegraphics[width=\columnwidth]{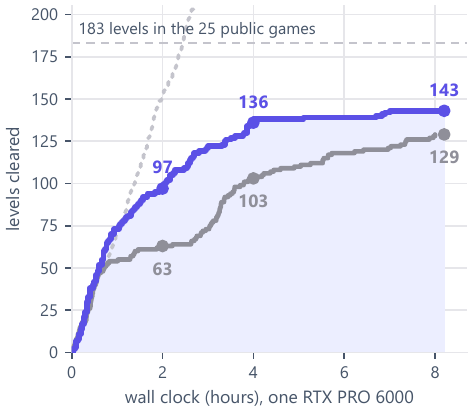}
\caption{\textbf{Cleared levels over wall-clock execution time on 25 public environments under identical hardware.} Indigo: \textsc{Sarbloh} (single pass, Run~A, briefing off). Solid grey: distinct levels cleared by the first Milestone winner (Duck harness) across four parallel passes; dotted grey: total levels summed across passes. \textsc{Sarbloh} clears 143 of 183 levels in 8.2 hours, surpassing the baseline's 129 distinct levels.}
\label{fig:curves}
\end{figure}

\paragraph{Hardware-matched comparative baseline.}
To isolate algorithmic contributions from compute resources, we evaluated \textsc{Sarbloh} directly against the first Milestone winner (the Duck harness \citep{duck2026}) on identical Kaggle hardware (one NVIDIA RTX PRO 6000) and identical Qwen3.8-Flash-Next model weights. While the baseline ran four parallel passes per game (100 concurrent streams), \textsc{Sarbloh} executed a single scheduled pass across the 25 public games. Crucially, this is Run~A, which ran with the briefing and map disabled and characterizes the action interface, serving loop, and scheduler (Appendix~\ref{app:telemetry}); in contrast, the official hidden-set score of 16.53 was produced with the full perception engine active (Figure~\ref{fig:leaderboard}). As shown in Figure~\ref{fig:curves}, \textsc{Sarbloh} consistently accelerated discovery, clearing 73 levels versus 54 at one hour, 97 versus 63 at two hours, and finishing at 143 of 183 levels in 8.2 hours versus 129 distinct levels for the baseline. This efficiency stems from our priority scheduler concentrating compute on game states with high marginal progress potential rather than partitioning resources uniformly.

\paragraph{Evaluation regime asymmetry and score contextualization.}
A critical distinction governs the comparison between offline competition submissions and frontier API benchmarks. Systems reporting near-ceiling scores on public games (Prime Agent, AVO, Arcgentica) operate across commercial API endpoints with unbounded wall-clock time, unrestricted internet access, and proprietary closed-weight backbones such as Claude Opus 5 and GPT-5.6 Sol. Furthermore, these evaluations execute on disclosed, publicly accessible environments where prompt templates and heuristic search budgets can be manually adapted without verifiable submission constraints. In contrast, \textsc{Sarbloh} operates within the official competition environment, where an immutable Docker container executes on a single offline GPU without external network access, constrained by a strict nine-hour wall-clock ceiling across the entire benchmark suite of undisclosed games. In this hardware-bounded regime, where every exploratory misstep incurs non-linear score degradation, achieving an official score of 16.53 represents an unprecedented result that surpasses the first Milestone winner by $13.7\times$, establishing that structured core-knowledge perception enables open-weight models to compete effectively without requiring unconstrained cloud resources.

\paragraph{Comparison with frontier model baselines and systems telemetry.}
On public environments, \textsc{Sarbloh} achieves an aggregate mean score of 72.5 across evaluation trials, outperforming proprietary frontier models evaluated via unconstrained APIs on 21 of 25 games and achieving 100\% resolution on 13 games (Figure~\ref{fig:pergame}). Under a single-pass regime, \textsc{Sarbloh} recorded a mean score of 59.4 in Run~A, briefing off (Table~\ref{tab:telemetry}), and 63.5 in Run~B, briefing on (Table~\ref{tab:onoff}), each exceeding the strongest frontier baseline (GPT-6.1 Sol at 54.8). 
In terms of runtime telemetry on the single Kaggle RTX PRO 6000 GPU, the append-only context preserved a 98.4\% prompt prefix-cache hit rate across 1.78\,B total prompt tokens, ensuring that prefill consumed only 1.3\% of busy execution time compared to 50.1\% for decode generation. Across 21,224 model calls, the runtime suffered zero server failures (Table~\ref{tab:telemetry}). The comparison of this briefing-off run with the nearest briefing-on run is in Appendix~\ref{app:telemetry} (Table~\ref{tab:onoff}). Runtime failure modes and ablated designs are detailed in Appendix~\ref{app:negative_results}.

\section{Related Work}
\label{sec:related}

\paragraph{Perceptual representations in interactive ARC agents.}
Prior ARC-AGI-3 architectures differ across planning strategies, context memory management, and external tooling, yet they share a uniform perceptual assumption: the visual grid is transmitted directly to the reasoning model as raw unparsed data, delegating spatial scene decomposition entirely to neural inference (Table~\ref{tab:field}). 
Specifically, the first Milestone winner (the Duck harness \citep{duck2026,arcprizemilestone1}) serializes the board as composite ASCII character matrices alongside rendered image frames. The second- and third-place Kaggle systems pass annotated screenshots to Gemma-4-31B, prompting the vision model to emit JSON action sequences directly \citep{arcprizemilestone1}. In the frontier API regime, VISTA \citep{vista2026} deploys a multimodal harness with lossless visual frame retention, while Arcgentica \citep{knutsen2026arcgentica} and Prime Agent \citep{karten2026prime} expose the environment within a persistent Python REPL, leaving the language model to write its own parsing scripts over raw coordinate arrays.

\paragraph{Core knowledge grounding and geometric invariants.}
Across all existing systems, the agent observes the world as an integer matrix, an ASCII string, or an unannotated visual raster. Consequently, fundamental spatial primitives, including 4-connected entity segmentation, topological containment, lattice-constrained reachability, and affine symmetries, are left to be reconstructed by the neural backbone from raw tokens. As our motivating ablations establish (Section~\ref{sec:motivation}), this unparsed representation forms the primary source of action inefficiency and hallucination. In contrast, \textsc{Sarbloh} computes geometric invariants deterministically on the host in $\mathcal{O}(HW)$ time, maintains persistent bipartite entity tracking across temporal steps, and explicitly decouples mathematical certainties from single-action testable affordance hypotheses.

\paragraph{Autonomous code generation and programmatic execution.}
The CodeAct paradigm \citep{wang2024codeact} demonstrated that expressing agent actions through executable programming primitives outperforms specialized JSON action schemas. Prime Agent \citep{karten2026prime} adapted this to ARC-AGI-3 by allowing models to execute Python code within an interactive environment. However, earlier harnesses decoupled model reasoning into disjoint tools for code simulation and action execution, forcing expensive context-switching roundtrips. \textsc{Sarbloh} unifies these operations within an asynchronous execution contract (\code{await play()}), allowing models to hypothesize, simulate, and commit multi-step actuation trajectories within a single reasoning turn.

\section{Limitations and Future Work}
\label{sec:future}

\paragraph{Limitations.}
Several architectural limitations remain. First, while affordance hypotheses are coupled with unit falsification tests, the current perception engine evaluates hypotheses independently at each frame without maintaining a fully unified Bayesian belief distribution across multi-step episodes. Second, although the agent plans with super-human efficiency once a level is understood, antecedent failure on complex perceptual transitions remains the primary source of score degradation. Finally, our current results rely on general-purpose open-weight models without task-specific policy fine-tuning.

\paragraph{Future work.}
Promising research avenues include:
\begin{enumerate}
    \item \textbf{Closed-Loop Belief Updating}: Formulating affordance induction as a dynamic partially observable belief state, where the empirical outcomes of unit falsification probes explicitly update prior belief weights across compositional levels.
    \item \textbf{Offline Trajectory Fine-Tuning}: Applying parameter-efficient fine-tuning (LoRA \citep{hu2021lora}) directly on our database of verified, high-efficiency execution trajectories, optimizing syntactic code generation and world model synthesis.
    \item \textbf{Sample-Efficient Reinforcement Learning}: Investigating offline reinforcement learning algorithms, such as Group Relative Policy Optimization (GRPO \citep{shao2024grpo}) and balanced policy clipping, to optimize planning depth and action parsimony within our fixed computational envelope.
\end{enumerate}

\section{Conclusion}
\label{sec:conclusion}

Autonomous reasoning in uninstructed interactive environments is fundamentally constrained by perception rather than downstream algorithmic planning. Under the non-linear action penalties of ARC-AGI-3, agents cannot afford to compensate for impoverished representations through exploratory trial and error. As our empirical investigations demonstrate, presenting environments as unparsed coordinate matrices induces perceptual misinduction, compelling language and vision backbones to squander their action budgets attempting to reconstruct topological adjacency and basic spatial mechanics.

To resolve this limitation, we introduced \textsc{Sarbloh}, an autonomous agent architecture organized around the Measured--Guessed Rule. By computing deterministic geometric invariants directly on the host in linear time, the perception engine extracts lattice reachability, topological enclosures, midpoint tethers, and dihedral symmetries as mathematical certainties that generalize unconditionally across novel environments. Simultaneously, the framework treats latent operational semantics not as brittle classifications, but as provisional inductive hypotheses paired with minimal unit-action verification probes. Grounded by this perceptual foundation, the agent models environment transitions through programmatic Python execution, dispatching multi-step actuation trajectories that halt dynamically upon encountering unexpected state transitions.

Our findings indicate that advancing fluid intelligence on novel benchmarks does not strictly necessitate larger parameter scales or unconstrained inference budgets. Instead, equipping reasoning models with structured, core-knowledge perceptual abstractions bridges the divide between visual observation and symbolic deduction, enabling open-weight architectures to explore with disciplined, sample-efficient physical agency.

\bibliography{references}

@article{arcagi3,
  title   = {{ARC-AGI-3}: A New Challenge for Frontier Agentic Intelligence},
  author  = {{ARC Prize Foundation}},
  journal = {arXiv preprint arXiv:2603.24621},
  year    = {2026}
}

@article{chollet2019measure,
  title   = {On the Measure of Intelligence},
  author  = {Chollet, Fran{\c{c}}ois},
  journal = {arXiv preprint arXiv:1911.01547},
  year    = {2019}
}

@article{spelke2007core,
  title   = {Core Knowledge},
  author  = {Spelke, Elizabeth S. and Kinzler, Katherine D.},
  journal = {Developmental Science},
  volume  = {10},
  number  = {1},
  pages   = {89--96},
  year    = {2007}
}

@article{karten2026prime,
  title   = {Prime Agent: A Self-Improving {RLM} Harness},
  author  = {Karten, Seth and Zhang, Alex L. and Thomas, Kevin and M{\"u}ller, Sebastian and Bakouch, Elie and others},
  journal = {arXiv preprint arXiv:2608.23552},
  year    = {2026}
}

@article{wang2024codeact,
  title   = {Executable Code Actions Elicit Better {LLM} Agents},
  author  = {Wang, Xingyao and Chen, Yangyi and Yuan, Lifan and Zhang, Yizhe and Li, Yunzhu and others},
  journal = {arXiv preprint arXiv:2402.01030},
  year    = {2024}
}

@article{rahmanzadehgervi2024blind,
  title   = {Vision Language Models Are Blind: Failing to Translate Detailed Visual Features into Words},
  author  = {Rahmanzadehgervi, Pooyan and Bolton, Logan and Taesiri, Mohammad Reza and Nguyen, Anh Totti},
  journal = {arXiv preprint arXiv:2407.06581},
  year    = {2024}
}

@article{zheng2023sglang,
  title   = {{SGLang}: Efficient Execution of Structured Language Model Programs},
  author  = {Zheng, Lianmin and Yin, Liangsheng and Xie, Zhiqiang and Sun, Chuyue and Huang, Jeff and others},
  journal = {arXiv preprint arXiv:2312.07104},
  year    = {2023}
}

@article{li2024eagle,
  title   = {{EAGLE}: Speculative Sampling Requires Rethinking Feature Uncertainty},
  author  = {Li, Yuhui and Wei, Fangyun and Zhang, Chao and Zhang, Hongyang},
  journal = {arXiv preprint arXiv:2401.15077},
  year    = {2024}
}

@article{yang2024gated,
  title   = {Gated Delta Networks: Improving {Mamba2} with Delta Rule},
  author  = {Yang, Songlin and Kautz, Jan and Hatamizadeh, Ali},
  journal = {arXiv preprint arXiv:2412.06464},
  year    = {2024}
}

@article{cheng2023autoround,
  title   = {Optimize Weight Rounding via Signed Gradient Descent for the Quantization of {LLMs}},
  author  = {Cheng, Wenhua and Zhang, Weiwei and Shen, Haihao and Cai, Yiyang and He, Xin and others},
  journal = {arXiv preprint arXiv:2309.05516},
  year    = {2023}
}

@article{hu2021lora,
  title   = {{LoRA}: Low-Rank Adaptation of Large Language Models},
  author  = {Hu, Edward J. and Shen, Yelong and Wallis, Phillip and Allen-Zhu, Zeyuan and Li, Yuanzhi and others},
  journal = {arXiv preprint arXiv:2106.09685},
  year    = {2021}
}

@article{shao2024grpo,
  title   = {{DeepSeekMath}: Pushing the Limits of Mathematical Reasoning in Open Language Models},
  author  = {Shao, Zhihong and Wang, Peiyi and Zhu, Qihao and Xu, Runxin and Song, Junxiao and others},
  journal = {arXiv preprint arXiv:2402.03300},
  year    = {2024}
}

@misc{duck2026,
  title        = {The Duck: {Tufa Labs}' {ARC-AGI-3} Harness, Milestone \#1 Winner},
  author       = {{Tufa Labs}},
  howpublished = {Kaggle notebook, \url{https://www.kaggle.com/code/jeroencottaar/tufa-labs-duck-harness-june-30-milestone-winner}},
  year         = {2026}
}

@misc{arcprizemilestone1,
  title        = {{ARC Prize} 2026: {ARC-AGI-3} Milestone \#1 Results},
  author       = {{ARC Prize Foundation}},
  howpublished = {\url{https://arcprize.org/blog/arc-prize-2026-milestone-1}},
  year         = {2026}
}

@misc{knutsen2026arcgentica,
  title        = {Arcgentica: {ARC-AGI-3} Agent Harness Built on the {Agentica} {SDK}},
  author       = {Knutsen, Samuel and Klein, Victoria},
  howpublished = {\url{https://github.com/symbolica-ai/ARC-AGI-3-Agents}},
  year         = {2026}
}

@misc{vista2026,
  title        = {{VISTA}: A Visual Harness for Reasoning in an Interactive World},
  author       = {Han, Qiushi and Hu, Keya and Qiu, Linlu and Wu, Cathy and He, Kaiming},
  howpublished = {\url{https://vista-research.github.io/}},
  year         = {2026}
}

@article{ullman1984visual,
  author    = {Shimon Ullman},
  title     = {Visual Routines},
  journal   = {Cognition},
  volume    = {18},
  number    = {1-3},
  pages     = {97--159},
  year      = {1984}
}

@book{marr1982vision,
  author    = {Marr, David},
  title     = {Vision: A Computational Investigation into the Human Representation and Processing of Visual Information},
  publisher = {W. H. Freeman},
  year      = {1982}
}

\clearpage
\appendix
\sloppy

\section{Notation and Priority Scheduler}
\label{app:notation}
\label{app:scheduler}

Table~\ref{tab:notation} collects the symbols used in the main text and in this appendix.

\begin{table}[h]
\centering
\small
\setlength{\tabcolsep}{4pt}
\begin{tabular}{@{}L{0.25\columnwidth}L{0.71\columnwidth}@{}}
\toprule
Symbol & Meaning \\
\midrule
\multicolumn{2}{@{}l}{\itshape\textcolor{muted}{Environment and score (Section~\ref{sec:prelim})}} \\
$g_t$ & grid observed at step $t$, in $\{0..15\}^{H\times W}$ \\
$H, W \le 64$ & grid height and width \\
$\mathcal{A}$ & action set, Equation~\eqref{eq:actions} \\
$n$;\ $l$;\ $k$ & levels in a game; level index; levels cleared \\
$h_l$;\ $a_l$ & human baseline actions; agent actions on level $l$ \\
$S_l$ & level score, Equation~\eqref{eq:level_score} \\
$E$ & game score, Equation~\eqref{eq:rhae} \\
\multicolumn{2}{@{}l}{\itshape\textcolor{muted}{Perception engine (Algorithm~\ref{alg:perceive})}} \\
$\mathcal{T}$ & terrain colours, each covering ${>}8\%$ of $HW$ \\
$L$ & layout: the things of a frame (objects, corridors, gauges, panels) \\
$\Lambda$;\ $\Lambda'$ & detected tile lattice; lattice of the previous frame \\
$D_4 \times \mathbb{Z}^+$ & the eight rotations and mirrorings of the square, with integer scaling \\
$F$ & measured facts (reach, enclosure, tethers, symmetries) \\
$e$;\ $\boldsymbol{\sigma}_e$ & a thing; its role scores \\
$\hat{\rho}_e$;\ $p_e$ & guessed role of $e$; confidence of that guess \\
\multicolumn{2}{@{}l}{\itshape\textcolor{muted}{Scheduler (Equation~\eqref{eq:priority})}} \\
$\pi$ & priority of a game for a GPU slot \\
$g_{\text{now}}$;\ $g_{\text{tail}}$ & value of the current level; of the levels after it \\
$m$;\ $a$ & reply tokens; actions on the current level \\
$h$;\ $m_0$;\ $a_0$ & 25 actions; 40k tokens; 115 actions \\
$\tau_{\text{tail}}$;\ $\phi$ & tail weight by levels left; end-of-run fade \\
\bottomrule
\end{tabular}
\caption{Notation.}
\label{tab:notation}
\end{table}

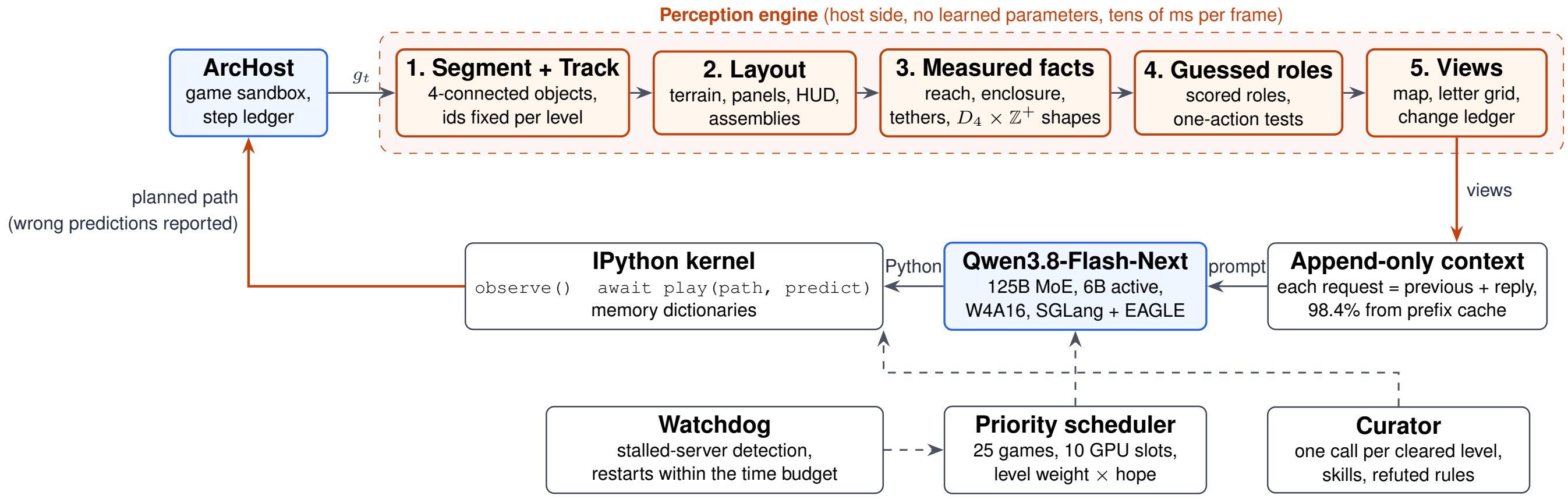
\begin{figure*}[t]
\centering
\resizebox{\textwidth}{!}{%
\begin{tikzpicture}[
  font=\sffamily\footnotesize, >={Stealth[length=2.2mm]},
  box/.style={draw=ink!80, rounded corners=3pt, fill=white, minimum height=7.5mm, inner sep=3.5pt, align=center, line width=0.55pt},
  hot/.style={box, fill=soft, draw=ours, line width=0.85pt},
  cool/.style={box, fill=softb, draw=base, line width=0.75pt},
  lab/.style={font=\sffamily\scriptsize, text=ink!90, align=center},
  flow/.style={->, draw=ink!75, line width=0.65pt},
  hotflow/.style={->, draw=ours, line width=0.95pt},
]
\node[cool, minimum width=21mm] (game) at (0,0) {\textbf{ArcHost}\\[-1pt]{\scriptsize game sandbox,}\\[-1pt]{\scriptsize step ledger}};
\node[hot, minimum width=27mm, right=9mm of game] (seg) {\textbf{1. Segment + Track}\\[-1pt]{\scriptsize 4-connected objects,}\\[-1pt]{\scriptsize ids fixed per level}};
\node[hot, minimum width=27mm, right=3mm of seg] (layout) {\textbf{2. Layout}\\[-1pt]{\scriptsize terrain, panels, HUD,}\\[-1pt]{\scriptsize assemblies}};
\node[hot, minimum width=27mm, right=3mm of layout] (measure) {\textbf{3. Measured facts}\\[-1pt]{\scriptsize reach, enclosure,}\\[-1pt]{\scriptsize tethers, $D_4 \times \mathbb{Z}^+$ shapes}};
\node[hot, minimum width=27mm, right=3mm of measure] (guess) {\textbf{4. Guessed roles}\\[-1pt]{\scriptsize scored roles,}\\[-1pt]{\scriptsize one-action tests}};
\node[hot, minimum width=24mm, right=3mm of guess] (render) {\textbf{5. Views}\\[-1pt]{\scriptsize map, letter grid,}\\[-1pt]{\scriptsize change ledger}};
\begin{scope}[on background layer]
\node[draw=ours, dashed, rounded corners=5pt, fill=ours!5, fit=(seg)(render), inner sep=6pt, label={[lab, text=ours, yshift=-1pt]above:\textbf{Perception engine} (host side, no learned parameters, tens of ms per frame)}] (pe) {};
\end{scope}
\node[box, minimum width=35mm, below=14mm of render.south east, anchor=north east] (ctx) {\textbf{Append-only context}\\[-1pt]{\scriptsize each request = previous + reply,}\\[-1pt]{\scriptsize 98.4\% from prefix cache}};
\node[cool, minimum width=35mm, left=8mm of ctx] (llm) {\textbf{Qwen3.8-Flash-Next}\\[-1pt]{\scriptsize 125B MoE, 6B active,}\\[-1pt]{\scriptsize W4A16, SGLang + EAGLE}};
\node[box, minimum width=45mm, left=8mm of llm] (kernel) {\textbf{IPython kernel}\\[-1pt]{\scriptsize \code{observe()} \quad \code{await play(path, predict)}}\\[-1pt]{\scriptsize memory dictionaries}};
\draw[flow] (game) -- node[lab, above]{$g_t$} (seg);
\draw[flow] (seg) -- (layout);
\draw[flow] (layout) -- (measure);
\draw[flow] (measure) -- (guess);
\draw[flow] (guess) -- (render);
\draw[hotflow] (render.south) -- node[lab, right]{views} (render.south |- ctx.north);
\draw[flow] (ctx) -- node[lab, above]{prompt} (llm);
\draw[flow] (llm) -- node[lab, above]{Python} (kernel);
\draw[hotflow] (kernel.west) -| node[lab, pos=0.75, left, align=right]{planned path\\(wrong predictions reported)} (game.south);
\node[box, minimum width=35mm, below=10mm of llm] (sched) {\textbf{Priority scheduler}\\[-1pt]{\scriptsize 25 games, 10 GPU slots,}\\[-1pt]{\scriptsize level weight $\times$ hope}};
\node[box, minimum width=35mm, right=8mm of sched] (cur) {\textbf{Curator}\\[-1pt]{\scriptsize one call per cleared level,}\\[-1pt]{\scriptsize skills, refuted rules}};
\node[box, minimum width=45mm, left=8mm of sched] (srv) {\textbf{Watchdog}\\[-1pt]{\scriptsize stalled-server detection,}\\[-1pt]{\scriptsize restarts within the time budget}};
\draw[flow, dashed] (sched) -- (llm);
\draw[flow, dashed] (cur.north) -- ++(0,4.5mm) -| (kernel.south east);
\draw[flow, dashed] (srv) -- (sched);
\end{tikzpicture}}
\caption{System topology of \textsc{Sarbloh}. The host-side perception engine extracts invariant relations and affordance hypotheses from each frame, passing views to an append-only context. The reasoning model generates Python search routines in a persistent kernel, committing multi-step paths via \code{await play()}. Twenty-five games share ten GPU execution slots under the utility-driven priority scheduler.}
\label{fig:overview}
\end{figure*}

\paragraph{Priority scheduler.}
Figure~\ref{fig:overview} shows the whole system. Twenty-five games share ten GPU slots, and the scheduler gives a free slot to the waiting game with the highest expected score gain. For a game that has cleared $k$ of its $n$ levels and has spent $m$ reply tokens and $a$ actions on level $k+1$ without clearing it, the priority is
\begin{align}
\pi &= \big(g_{\text{now}} + g_{\text{tail}}\big)\cdot \underbrace{2^{-\left((m/m_0)^{2} + (a/a_0)^{2}\right)}}_{\text{hope}}, \label{eq:priority}\\
g_{\text{now}} &= \frac{k+1}{n(n+1)/2}\Big(\frac{h}{h+a}\Big)^{2}, \quad g_{\text{tail}} = \frac{\tau_{\text{tail}}(n-k-1)}{55}\,\phi. \nonumber
\end{align}
$g_{\text{now}}$ is the level's weight in Equation~\eqref{eq:rhae}, discounted quadratically by the actions already spent on it; $g_{\text{tail}}$ is the value of the later levels that clearing it unlocks. We set $h = 25$ actions (a typical human level), $m_0 = 40$k tokens and $a_0 = 115$ actions. $\tau_{\text{tail}}(j)$ is 8, 7, 5 or 0 when $j \ge 3$, 2, 1 or 0 levels remain after the current one, on the 55-weight scale of a 10-level game, so short games do not starve long ones. $\phi \in [0, 1]$ falls linearly to zero over the last 20\% of the run, when there is no time left to reach later levels. A game that has never held a slot goes first, and a game keeps its slot for 20 requests before the scheduler looks again. The constants were set by hand, without a search. Equation~\eqref{eq:priority} is the function \code{priority} in \code{Sarbloh/harness/scheduler.py}.

\subsection{Hyperparameters}
\label{app:config}

Table~\ref{tab:hyperparams} lists the settings of the two full public-game runs of Appendix~\ref{app:telemetry}, read from the configuration file each run saved with its results and from its server log. The two runs differ only where a cell gives two values.

\begin{table}[h]
\centering
\footnotesize
\setlength{\tabcolsep}{4pt}
\begin{tabular}{@{}L{0.52\columnwidth}L{0.44\columnwidth}@{}}
\toprule
Parameter & Value \\
\midrule
Games / GPU slots & 25 / 10 \\
Scheduler quantum & 20 model calls \\
Action cap per game & 2,000 actions \\
Actions per \code{play()} path & at most 100 \\
Reply cap per request / requests per turn & 12k tokens / 12 \\
Thinking without a move before the turn yields & 16,384 tokens \\
Context drain trigger / target & 118k / 55k (Run A), 70k (Run B) \\
Context window & 131,072 tokens \\
Map scale & 10 pixels per grid cell \\
Briefing lines / short-view text cap & 20 lines / 600 tokens \\
Memory caps: goal / plan / hypotheses / findings & 150 / 200 / 400 / 300 tokens \\
Temperature / top-$p$ / top-$k$ & 0.7 / 0.95 / 20 \\
Serving & SGLang, W4A16 (AutoRound), FP8 KV cache \\
Speculative decoding & EAGLE with a multi-token-prediction drafter, 3 steps, 4 draft tokens \\
\bottomrule
\end{tabular}
\caption{Settings of Run~A (briefing off) and Run~B (briefing on).}
\label{tab:hyperparams}
\end{table}

\section{Perception Lab Protocols and Full Briefings}
\label{app:lab}

This appendix gives the protocol of the perception lab, the eight presentations it compared, and the full briefings for the two lab frames. The results are in Section~\ref{sec:lab_results} (Table~\ref{tab:lab}, Figure~\ref{fig:lab} and Table~\ref{tab:holdout}).

\subsection{Setup and Answer Key}
\label{sec:lab}

\paragraph{Shared front end.}
All experimental arms (X1--X8) share an identical front end to isolate representational formatting from extraction noise. A raw frame is transformed into an exact $64{\times}64$ categorical color grid by fitting cell pitch and origin from chromatic boundary gradients. The grid is segmented into background terrain (modal color coverage exceeding 8\%), four-connected monochromatic entities categorized by morphology (compact objects, icons, dotted structures, thin connectors, or bars), bounding panels, and peripheral heads-up display (HUD) elements within three cells of the screen boundary.

\paragraph{Baseline encoder.}
The baseline is our legacy text-only encoder (\code{arc\_eye}). It extracts the fitted lattice, assigns a fixed palette symbol per cell, performs connected component segmentation with hole detection, computes dihedral shape hashes, and identifies bounding boxes alongside inter-frame coordinate differences. It emits an exhaustive text dump requiring approximately 3,500 tokens per frame without visual images.

\paragraph{Answer key and scoring.}
The evaluation suite comprises the opening levels of \code{ls20} (arrow-key lattice navigation) and \code{r11l} (continuous coordinate click interactions). Before any arm was run, an answer key of seven items was written for each frame:
\begin{description}
    \item[K1] the controllable entity;
    \item[K2] the goal;
    \item[K3] the link or carried object;
    \item[K4] the modifier or hazard;
    \item[K5] the status gauges;
    \item[K6] the winning plan;
    \item[K7] the terrain boundaries.
\end{description}
Each item scored 2 for a clear and correct description, 1 for a partial one, 0 if it was missing and $-1$ if it was misleading, so each game has a maximum of 14 points and the two together 28. An intuition rating from 1 to 5 and the prompt token cost were recorded alongside.

\begin{figure*}[t]
\centering
\includegraphics[width=\textwidth]{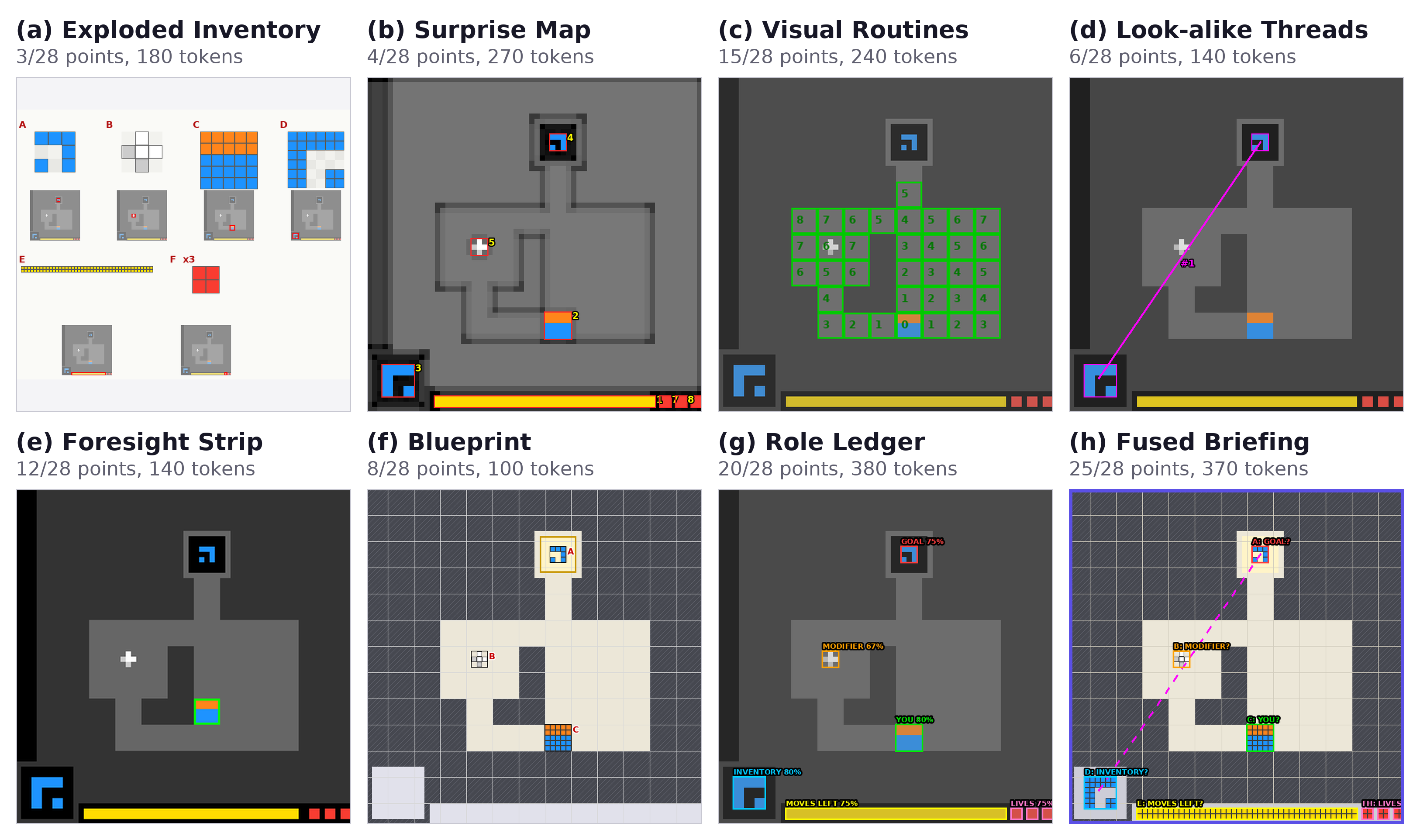}
\caption{\textbf{Visual presentations evaluated in the Perception Lab on \code{ls20}.} (a) Exploded Inventory separates unique entity morphologies. (b) Surprise Map highlights unpredicted local textures. (c) Visual Routines computes lattice BFS step counts to candidate targets. (d) Look-alike Threads connects shapes matching under rotation, reflection, or integer scaling. (e) Foresight Strip simulates counterfactual forward states. (f) Blueprint redraws terrain and lattices as a clean architectural schematic. (g) Role Ledger annotates candidate affordance hypotheses. (h) Fused Briefing combines the blueprint map with measured invariants and testable role hypotheses.}
\label{fig:labgallery}
\end{figure*}

\subsection{Candidate Presentations}
\label{sec:lab-exps}

\begin{description}
    \item[X1 Exploded Inventory.] Entities are de-duplicated by exact shape grids, magnified to a standard display scale, arranged spatially with letter tags, and annotated with multiplicity counts ($\times N$). A miniature thumbnail outlines original entity positions. Thin connectors are cataloged by endpoint coordinates.
    \item[X2 Surprise Map.] Calculates cell-level information surprise as $-\log_2 p$ based on $3{\times}3$ neighborhood pattern distributions. Periodic floor tiles and boundary walls score near zero bits. Entities are ranked by total bits, preserving full saturation only on high-surprise components while fading predictable background textures.
    \item[X3 Visual Routines.] Executes serial geometric routines based on \citet{ullman1984visual}. \textbf{Reach} performs breadth-first search across detected lattice steps, reporting path lengths from the likely agent to all visible targets. \textbf{Fit} correlates hollow enclosing boundaries with objects of identical dimensions. \textbf{Trace} tracks single-pixel connectors to their endpoints, identifying entities suspended at the midpoint of dual anchors. \textbf{Align} catalogs entities sharing identical row or column spans.
    \item[X4 Look-alike Threads.] Compares binary entity shapes under all eight dihedral symmetries of the square alongside integer scaling factors ($D_4 \times \mathbb{Z}^+$). Matched pairs are classified as \textsc{exact} (identical copies), \textsc{turned} (matched under rotation or reflection), \textsc{near} (differing by at most three cells), or \textsc{state} (identical geometry with modified color palettes). Visual threads connect matched entities across the frame.
    \item[X5 Foresight Strip.] Projects prospective states for legal actions using hardcoded physical priors: agent sprites translate by their bounding dimensions until striking walls, edge gauges decrement by unit length per action, and midpoint objects track anchor displacements.
    \item[X6 Blueprint.] Re-renders the scene as an architectural schematic following \citet{marr1982vision}: terrain rendered in soft tones, structural boundaries hatched, tile lattices faintly gridded, and HUD elements isolated into a dedicated side panel.
    \item[X7 Role Ledger.] Scores non-terrain entities against eight standard gameplay roles (\textsc{You}, \textsc{Goal}, \textsc{Moves Left}, \textsc{Lives}, \textsc{Inventory}, \textsc{Modifier}, \textsc{Handle}, \textsc{Carried Thing}) using evidence heuristics, printing each top candidate alongside an empirical unit-action verification test.
    \item[X8 Fused Briefing.] Synthesizes the architectural blueprint canvas of X6, the relational threads and midpoint indicators of X3--X4, and a two-tiered text briefing explicitly separating \textsc{Measured} geometric certainties from \textsc{Guessed} affordance hypotheses paired with unit-action verification tests.
\end{description}

Figure~\ref{fig:labgallery} provides a visual comparison of each candidate presentation evaluated on \code{ls20}.

\subsection{Full Structural Briefings}
\label{app:briefings}

The two boxes below give the briefing text the perception engine wrote, word for word, for the opening frames of \code{ls20} (Figure~\ref{fig:views}) and \code{r11l}. The object ids are the same as those on the map and in the kernel's object table (\code{observe().objects}).

\begin{briefing}{\code{ls20}, opening frame (arrow keys, tile grid)}
BRIEFING (arrow keys)
MEASURED (certain):
- Floor is a 5x5 tile grid.
- obj 4 (blue 6-cell shape): sits inside a black box that is NOT plain floor; the floor reaches the box's bottom side in 5 steps (the only way in)
- obj 6 (white/light grey plus): reachable in 6 steps
- obj 4 (blue 6-cell shape) is directly above obj 10 (blue/orange square) (shared columns 35-37)
- obj 12 (blue irregular shape (24 cells)) (drawn 2x bigger), mirrored left-right OR turned 90 deg clockwise, becomes exactly obj 4 (blue 6-cell shape)
- 3 identical red square (2x2): obj 16, 17, 18 -> a count of something (lives, items, tiles)
- obj 6 (white/light grey plus) has a symmetric shape but lopsided colours (marks a direction/state)
- obj 10 (blue/orange square) has a symmetric shape but lopsided colours (marks a direction/state)
GUESSES (check with one action):
- obj 10 (blue/orange square) = YOU (80%) - test: press one arrow key: it should move one tile and the gauge should drop
- obj 12 (blue irregular shape (24 cells)) = INVENTORY (80%) - test: watch it after touching the modifier
- obj 4 (blue 6-cell shape) = GOAL (75%) - test: get the right thing into it and see if the level ends
- obj 15 (yellow bar) = MOVES LEFT (75%) - test: take any action and check it shrinks by one unit
- obj 16, 17, 18 (red square) = LIVES (75%) - test: watch it when the gauge runs out or you fail
- obj 6 (white/light grey plus) = MODIFIER (67%) - test: step onto it, then check whether the inventory icon changed
\end{briefing}

\begin{briefing}{\code{r11l}, opening frame (clicks, no tile grid)}
BRIEFING (clicks)
MEASURED (certain):
- no tile grid -> free movement. Open space = black (81% of screen).
- all 4 main objects share one open space; straight paths between them are ALL CLEAR of rock
- the grey areas are not open floor (likely walls)
- obj 4 (purple dotted ring) encloses an EMPTY 5x5 space -- exactly the size of: obj 16 (white/purple diamond), obj 25 (purple/magenta disc), obj 36 (dark grey/purple diamond)
- a light grey line connects obj 16 (white/purple diamond) to obj 25 (purple/magenta disc)
- a light grey line connects obj 25 (purple/magenta disc) to obj 36 (dark grey/purple diamond)
- obj 25 (purple/magenta disc) sits at the MIDPOINT of obj 16 (white/purple diamond) and obj 36 (dark grey/purple diamond) (error 0.5 cells) -> it is held between them; moving either end should move it to the new midpoint
- obj 16, 36 (white/purple diamond; dark grey/purple diamond) have the same shape in different colours (white/purple vs dark grey/purple)
GUESSES (check with one action):
- obj 4 (purple dotted ring) = GOAL (80%) - test: get the right thing into it and see if the level ends
- obj 16, 36 (white/purple diamond; dark grey/purple diamond) = HANDLE (click) (80%) - test: click an empty spot once: the selected handle should jump there
- obj 1 (white bar) = MOVES LEFT (75%) - test: take any action and check it shrinks by one unit
- obj 25 (purple/magenta disc) = CARRIED THING (75%) - test: move one handle: it should slide to the new midpoint
\end{briefing}

\subsection{The Perception Pipeline on Four More Games}
\label{app:pipelines}

\begin{figure*}[tp]
\centering
\includegraphics[width=\textwidth]{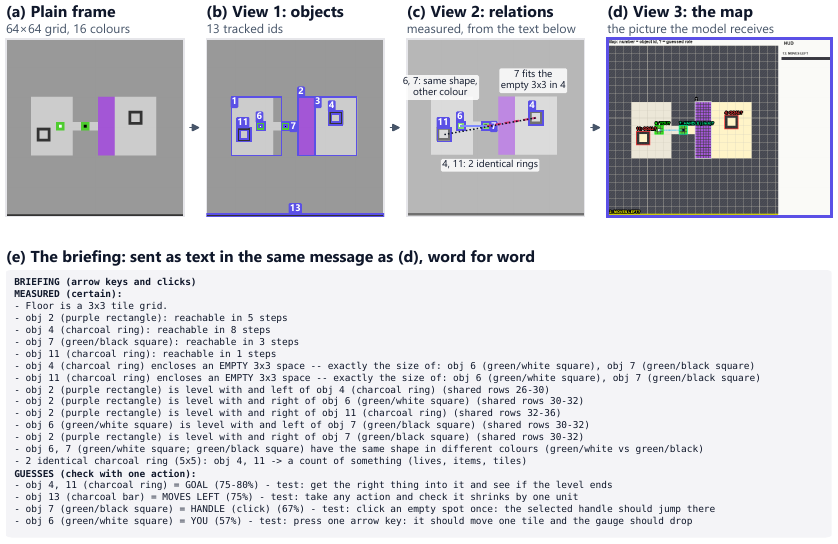}
\caption{The perception pipeline on the opening frame of \code{ka59} (arrow keys and clicks, $3{\times}3$ tile grid). (c)~Rings 4 and 11 are identical; ring 4 encloses an empty $3{\times}3$ that exactly fits square 7 (and square 6); squares 6 and 7 have the same shape in different colours. Panels as in Figure~\ref{fig:views}; the model receives (d) and (e).}
\label{fig:pipe-ka59}
\end{figure*}

\begin{figure*}[tp]
\centering
\includegraphics[width=\textwidth]{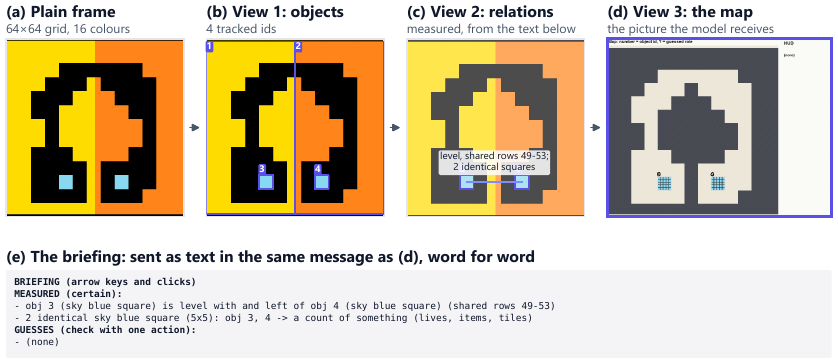}
\caption{The perception pipeline on the opening frame of \code{m0r0} (arrow keys and clicks). The engine finds four objects and measures two facts: squares 3 and 4 are identical and share rows 49--53. It makes no guesses. Panels as in Figure~\ref{fig:views}.}
\label{fig:pipe-m0r0}
\end{figure*}

\begin{figure*}[tp]
\centering
\includegraphics[width=\textwidth]{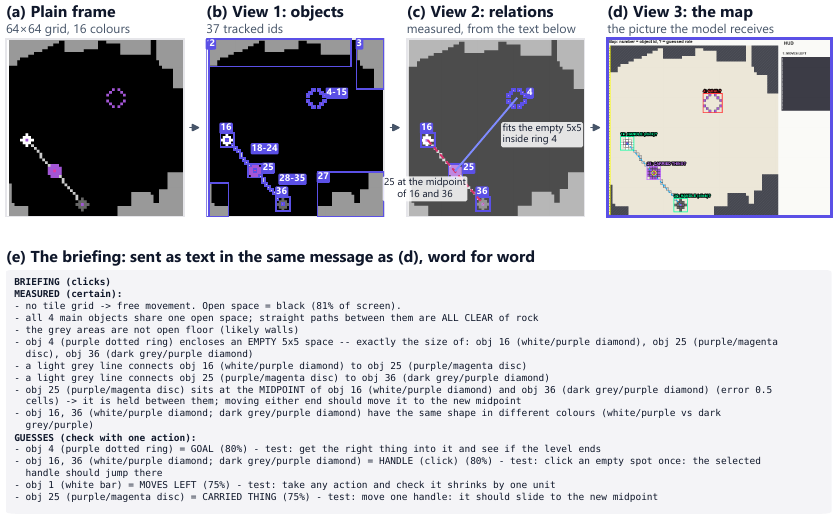}
\caption{The perception pipeline on the opening frame of \code{r11l} (clicks, no tile grid). (b)~The dotted ring (4--15) and the two connecting lines (18--24, 28--35) are made of one-cell parts and are tagged once each. (c)~Disc 25 sits at the midpoint of handles 16 and 36, and fits the empty $5{\times}5$ inside ring 4. Panels as in Figure~\ref{fig:views}.}
\label{fig:pipe-r11l}
\end{figure*}

\begin{figure*}[tp]
\centering
\includegraphics[width=\textwidth]{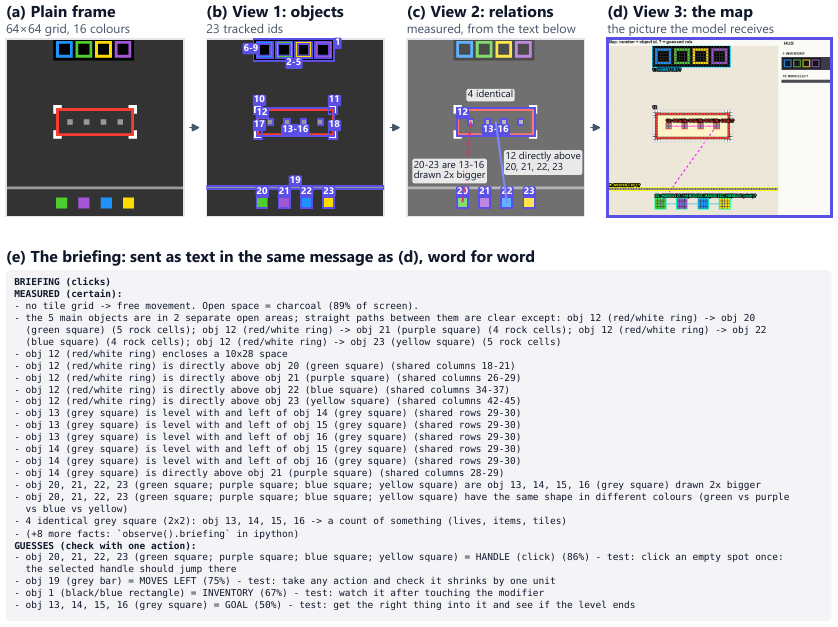}
\caption{The perception pipeline on the opening frame of \code{sb26} (clicks only). (c)~Squares 20--23 are the grey squares 13--16 drawn 2$\times$ bigger, and ring 12 sits directly above each of them. (e)~The briefing reaches its line limit; the last MEASURED line says eight more facts can be read from \code{observe().briefing}. Panels as in Figure~\ref{fig:views}.}
\label{fig:pipe-sb26}
\end{figure*}

Figures~\ref{fig:pipe-ka59}--\ref{fig:pipe-sb26} repeat Figure~\ref{fig:views} for the opening frames of \code{ka59}, \code{m0r0}, \code{r11l} and \code{sb26}. They were made in the same way: \code{make\_perception\_figure.py} runs the agent's perception code on the first frame of each local game, and panels (b) and (c) are drawn from its object table and briefing. Each line in (c) stands for one MEASURED line of the briefing in (e). Where an object is made of one-cell parts (the dotted ring and the connecting lines in \code{r11l}) or repeats (the four grey squares in \code{sb26}), the parts are outlined and the group gets one tag. The four games test different parts of the engine. \code{ka59} has a tile grid, and two rings each enclose an empty $3{\times}3$ that fits a movable square. \code{m0r0} is sparse: four objects, two measured facts and no guesses, so the engine does not invent roles it has no evidence for. \code{r11l} is driven by clicks, and its disc sits at the midpoint of two connected handles. \code{sb26} is the densest: its briefing reaches the line limit, and the remaining eight facts are left to \code{observe().briefing} in the kernel, as the last MEASURED line says.

\section{Negative Results, Ablations, and Runtime Boundary Conditions}
\label{app:negative_results}

This appendix covers the designs that failed or made the agent unstable in live runs, the runtime failure modes, and the two full runs nearest to a perception ablation.

\subsection{Perceptual Edge Cases and Line Budget Quotas}

\paragraph{Lattice detection instability from peripheral HUD elements.}
In early iterations, the host-side lattice detection module fitted tile dimensions over all non-terrain pixels. In \code{ls20}, when the player moved, an off-grid gauge within the status box introduced irregular coordinate intervals, causing the fitted $5{\times}5$ grid to collapse. Similarly, in \code{ka59}, touching a secondary sprite altered the detected lattice periodicity. Each collapse temporarily deleted all reachability invariants from context, causing the agent to vacillate between conflicting spatial hypotheses. We resolved this by explicitly excluding peripheral HUD regions from the lattice fitting routine and caching the initial valid lattice configuration for the remainder of the level.

\begin{figure*}[t]
\centering
\input{figures/briefing_budget.tex}
\caption{\textbf{The briefing line budget on the first frame of \code{tr87}}, both boxes built by the agent's own \code{Briefing} code (object descriptions shortened; \code{r}/\code{c} = shared rows/columns). The mark at the left of each line gives its kind: \bmark{base}enclosure, \bmark{muted}alignment, \bmark{lavender}look-alike, \bmark{ours}guess. (a)~One cap of 30 lines over all facts in the engine's order: 23 of the 30 lines are alignments, and the look-alike pair and both guesses are cut. (b)~The policy the agent runs, which shares the lines between fact kinds and keeps up to a third for guesses: it shows the look-alike pair and the two guesses and counts the 40 facts left out (``\code{+40 more facts}'').}
\label{fig:budget}
\end{figure*}

\paragraph{Combinatorial relation crowding.}
In scenes with many objects, pairwise relations grow quadratically. The first frame of \code{tr87} yields 60 alignment pairs, 7 enclosures and 1 symmetry, and \code{bp35} yields 792 alignment pairs. Under one global line cap the alignments fill the briefing and push out the symmetry and the role guesses (Figure~\ref{fig:budget}a). The deployed policy shares the lines between fact kinds, keeps up to a third of them for guesses and reports the rest as a count (Figure~\ref{fig:budget}b).

\subsection{Rejected Architectural Designs}

\paragraph{Sparse delta state encoding.}
Transmitting only differential coordinate changes between frames substantially reduces prompt token consumption. However, in ARC-AGI-3 environments where local actions trigger distant causal effects (e.g., activating a floor plate updating an icon across the display), sparse updates conceal non-local covariance. In development traces, models receiving delta updates consistently concluded that actions were non-functional when localized pixels remained static, leading to catastrophic action loops. Full-frame visual representations remain essential for maintaining causal coherence.

\paragraph{Disjoint tool spaces for reasoning and actuation.}
Providing separate tool interfaces for code execution and environment stepping caused models to fragment reasoning trajectories. The agent formulated world models in code, exited the REPL to dispatch individual actions, and re-entered code execution to reconstruct state variables, multiplying inference latency. Merging action commitment directly into the Python kernel via \code{await play()} was the single intervention responsible for advancing the hidden-set score from 5.14 to 5.65.

\paragraph{Semantic history comment stubbing.}
To conserve context window capacity, we tested replacing executed historical code cells with brief descriptive comments. In the two test runs that had stubbing switched on, reasoning backbones degenerated into copying comment strings rather than generating executable Python code (occurring in 97.7\% of turns and clearing 0 of 97 levels). Smaller open-weight models heavily imitate local context formatting; historical code must remain intact as functional syntax.

\paragraph{Autonomous subagent hierarchies.}
Providing the model with programmatic subagent delegation primitives (following the Prime Agent specification) yielded zero productive invocations across all benchmark runs. Open-weight 6B-active models failed to effectively coordinate subordinate contexts, adding prompt complexity without empirical utility.

\paragraph{Self-imitation fine-tuning on cleared trajectories.}
We fine-tuned Qwen3.8-Flash-Next on successful execution traces using low-rank adaptation (LoRA). While training cross-entropy loss decreased from 11\% to 6\%, out-of-distribution benchmark performance degraded significantly. The fine-tuned weights exhibited rigid memorization of specific coordinate paths rather than generalized spatial reasoning, consuming over 80\% of our training compute budget without positive yield.

\subsection{Interactive Runtime Failure Modes}

\paragraph{Degenerate holding loops at the action cap.}
In Run~A (briefing off), four games (\code{lf52}, \code{s5i5}, \code{vc33}, \code{re86}) stopped 2 to 15 actions short of their 2,000-action cap. The harness kept asking them to continue, and the model answered ``\code{Holding...}'' about 14,000 times, which used 1.24\,B of the run's 1.78\,B prompt tokens. The stalls began after these games had cleared every level they would clear, so they cost wall clock, not score. The harness has a per-game turn limit (400 by default), but both full runs set it to $10^6$, so no run reported here was protected by it; whether the limit removes the stall without costing levels is untested.

\subsection{Runtime Telemetry and the Nearest Briefing Ablation}
\label{app:telemetry}

\paragraph{The run was decode-bound.}
Table~\ref{tab:telemetry} gives the telemetry of Run~A, briefing off (8.2\,h, 25 games, one RTX PRO 6000), from the run's speed report and server log. Because 98.4\% of the prompt tokens came from the prefix cache, prefill took 1.3\% of busy time and decoding 50.1\%. The server answered 21,224 model calls without a failure.

\begin{table}[h]
\centering
\small
\begin{tabular}{@{}L{0.66\columnwidth}r@{}}
\toprule
Run A: 25 games, one GPU, 8.2\,h & Value \\
\midrule
Prompt tokens served from the prefix cache & \textbf{98.4\%} of 1.78\,B \\
Decode speed per request & 76 tok/s \\
Busy time: decode / prefill & 50.1\% / 1.3\% \\
Model calls / server failures & 21,224 / 0 \\
\bottomrule
\end{tabular}
\caption{Systems telemetry of Run~A (briefing off).}
\label{tab:telemetry}
\end{table}

\paragraph{The 143-level run sent no briefing.}
Run~A was made while the briefing and map were switched off: the model saw the plain $640{\times}640$ frame and the letter grid. Its numbers describe the action interface, the serving loop and the scheduler, not the perception engine.

\paragraph{The closest briefing-on run cleared levels more slowly but more efficiently.}
Run~B had the briefing and map on, together with new guardrails and a stuck-level review, and ran for 5.4\,h (Table~\ref{tab:onoff}). It cleared fewer levels in the first two hours (51 against 74 at 1\,h) and drew level at about 4\,h, while its median action ratio on cleared levels fell from 0.77 to 0.68 and its mean public score rose from 59.4 to 63.5. The runs differ in more than the briefing and there is one run per arm, so we report this as consistent with fewer wasted actions once a level is understood, not as a measured effect of perception.

\begin{table}[h]
\centering
\small
\setlength{\tabcolsep}{4pt}
\begin{tabular}{@{}lrr@{}}
\toprule
25 public games, one GPU & Run A & Run B \\
Logged (2026) & 23 Oct & 27 Oct \\
\midrule
\multicolumn{3}{@{}l}{\itshape\textcolor{muted}{Configuration}} \\
Briefing and map & off & on \\
Extra guardrails, stuck review & -- & \checkmark \\
Context drain target & 55k & 70k \\
Wall clock & 8.2\,h & 5.4\,h \\
\multicolumn{3}{@{}l}{\itshape\textcolor{muted}{Outcome}} \\
Levels cleared at 1\,h / 2\,h / 4\,h & 74 / 99 / 138 & 51 / 90 / 131 \\
Levels cleared at the end & 143 & 138 \\
Median $a_l/h_l$ on cleared levels & 0.77 & \textbf{0.68} \\
Mean public score & 59.4 & \textbf{63.5} \\
Prompt tokens from cache & 98.4\% & 93.3\% \\
\bottomrule
\end{tabular}
\caption{The two full runs nearest to a perception ablation. They differ in more than the briefing, so the comparison is suggestive only. Level counts are from each run's own log of clear times.}
\label{tab:onoff}
\end{table}

\section{Open-Source Artifacts and Reproducibility}
\label{app:reproduce}

\paragraph{Licence.}
All code we wrote, the model-serving scripts, the evaluation harness and the experiment logs are public under the MIT-0 licence, a public-domain-equivalent licence as the competition rules require, at \url{https://github.com/Harsidak/SARBLOH}. Third-party components keep their own licences and are listed in the repository's \code{NOTICE} file.

\paragraph{Kaggle notebook and submission.}
The official scored benchmark evaluation was submitted directly from the public Kaggle competition notebook available at \url{https://www.kaggle.com/code/banwait13/sarbloh-experimentation}.

\paragraph{Run log.}
The paper names its two full public runs by configuration. In the experiment log, Run~A (briefing off) is the run of 23~October 2026 and Run~B (briefing on) the run of 27~October 2026, as in Table~\ref{tab:onoff}.

\paragraph{Scripts and evaluation reproduction.}
The figures, and the numbers they plot, are produced by reproducible scripts in the released repository, under \href{https://github.com/Harsidak/SARBLOH/tree/main/Writing/scripts}{\code{Writing/scripts/}} (with input tables in \href{https://github.com/Harsidak/SARBLOH/tree/main/Writing/data}{\code{Writing/data/}}):
\begin{itemize}
    \item \code{collect\_data.py}: parses submission logs, per-game score tables, and perception rubrics into \code{Writing/data/}.
    \item \code{extract\_curves.py}: extracts levels cleared over wall-clock time from raw run logs.
    \item \code{error\_budget.py}: splits points lost in a run into unreached levels, attempted levels, and action overhead under official RHAE weights.
    \item \code{make\_perception\_figure.py}: runs the perception engine on real game frames to render objects, maps, and briefings.
    \item \code{make\_briefing\_budget.py}: generates the two \code{tr87} briefings of Figure~\ref{fig:budget} using the agent's \code{Briefing} module.
    \item \code{make\_method\_figures.py}: renders Method and Results figures from run telemetry.
    \item \code{plot\_results.py}: renders benchmark curves and evaluation figures as vector PDF and PNG.
\end{itemize}

\end{document}